\documentclass[final,5p,times,twocolumn,authoryear]{elsarticle}

\usepackage{amssymb}
\usepackage{lineno}
\usepackage{hyperref}

\usepackage{amsmath}
\usepackage{bm}
\usepackage{xcolor}
\usepackage{booktabs}
\usepackage{multirow}

\journal{Neural Networks}

\begin{document}
\begin{frontmatter}



\title{MAProtoNet: A Multi-scale Attentive Interpretable Prototypical Part Network for 3D Magnetic Resonance Imaging Brain Tumor Classification \tnoteref{}}

% \cortext[cor1]: *
% \cortext[cor1]: **

% \author{Binghua LI \corref{cor1} \fnref{aff1}}
% \ead{s237857s@st.go.tuat.ac.jp}

\author[aff1,aff2]{Binghua Li}
\ead{nkvhua@outlook.com}

\author[aff3]{Jie Mao}

\author[aff3,aff4,aff5]{Zhe Sun}

\author[aff2]{Chao Li}

\author[aff2,aff1]{Qibin Zhao}

\author[aff1,aff2]{Toshihisa Tanaka \corref{cor1}}

% define footnote
\cortext[cor1]{corresponding author}
\affiliation[aff1]{organization={Tokyo University of Agriculture and Technology (TUAT)},
                   city={Tokyo},
                   country={Japan}
        }

\affiliation[aff2]{organization={RIKEN Center for Advanced Intelligence Project (RIKEN AIP)},
                   city={Tokyo},
                   country={Japan}
        }

\affiliation[aff3]{organization={Graduate School of Medicine, Juntendo University},
                   city={Tokyo},
                   country={Japan}
        }

\affiliation[aff4]{organization={Faculty of Health Data Science, Juntendo University},
                   city={Tokyo},
                   country={Japan}
        }
 
\affiliation[aff5]{organization={Image Processing Research Team, RIKEN},
                   city={Tokyo},
                   country={Japan}
        }
   
% \affiliation{organization={},

\begin{abstract}
\label{abstract}
Automated diagnosis with artificial intelligence has emerged as a promising area in the realm of medical imaging, while the interpretability of the introduced deep neural networks still remains an urgent concern. Although contemporary works, such as XProtoNet \citep{kim_xprotonet_2021} and MProtoNet \citep{wei_mprotonet_2024}, has sought to design interpretable prediction models for the issue, the localization precision of their resulting attribution maps can be further improved. 
To this end, we propose a \textbf{\underline{M}}ulti-scale \textbf{\underline{A}}ttentive \textbf{\underline{Proto}}typical part \textbf{\underline{Net}}work, termed MAProtoNet, to provide more precise maps for attribution. Specifically, we introduce a concise multi-scale module to merge attentive features from quadruplet attention layers, and produces attribution maps. The proposed quadruplet attention layers can enhance the existing online class activation mapping loss via capturing interactions between the spatial and channel dimension, while the multi-scale module then fuses both fine-grained and coarse-grained information for precise maps generation. We also apply a novel multi-scale mapping loss for supervision on the proposed multi-scale module. 
Compared to existing interpretable prototypical part networks in medical imaging, MAProtoNet can achieve state-of-the-art performance in localization on brain tumor segmentation (BraTS) datasets, resulting in approximately 4\% overall improvement on activation precision score (with a best score of 85.8\%), without using additional annotated labels of segmentation. 
Our code will be released in \href{https://github.com/TUAT-Novice/maprotonet}{https://github.com/TUAT-Novice/maprotonet}.

\end{abstract}


% \begin{highlights}
% \end{highlights}


\end{frontmatter}

\section{Introduction}
\label{introduction}
% Part 1. 引入
% 1.1. 阐释深度学习 + 医学图像
% AI for medicine is important
Over the past decade, artificial intelligence (AI), particularly in the realm of deep learning (DL) and computer vision (CV), has witnessed substantial  progress in architecture design \citep{vaswani_attention_2017,he_deep_2016, ronneberger_u-net_2015,ho_denoising_2020} and training strategy of deep neural network \citep{ioffe_batch_2015,srivastava_dropout_nodate,nair_rectified_nodate}. These artful architectures and ingenious training strategies endow convolution neural networks (CNNs) and transformers with unmatched potency than ever before \citep{dosovitskiy_image_2021,wang_convolutional_2022,cheng_masked-attention_2022}. These breakthroughs, in turn, showcase the promising prospects of CNNs and transformers in medical imaging practices, encompassing areas such as diagnosis, object detection, semantic segmentation and so on \citep{shastry_cancer_2022}. 
% 1.2. 引出可解释性
% 拓展
Yet, as the applied models becoming increasingly deep and complex, the concerns regarding their interpretablility are also growing, especially in the field of medical imaging where decisions are high-stake and high-consequence \citep{leming_challenges_2023,patricio_explainable_2024}.
% a. Challenges of implementing computer-aided diagnostic models for neuroimages in a clinical setting
% b. Explainable Deep Learning Methods in Medical Image Classification: A Survey
In DL, models are usually referred to as black-boxes since their inner states are not as straightforward or easy to interpret as models such as decision trees. However, as for medical practices, “transparency” of the models is quite essential \citep{salahuddin_transparency_2022}. 
% Transparency of deep neural networks for medical image analysis: A review of interpretability methods
Furthermore, model interpretability can not only present reasoning process while aiding clinicians in their decision-making but is also, in many cases, aligns with ethical and even legal requirements (e.g. the GDPR in European Union \citep{temme_algorithms_2017}). 
% Algorithms and transparency in view of the new general data protection regulation (e.g. 法律)
% 1.3. 总结
Consequently, despite considerable advancements in DL applications for medical imaging, the ongoing research on model interpretability remains necessary.

% Part 2. 图像可解释性
Interpretability, however, can be achieved through various means. Models deemed interpretable include those capable of providing auxiliary information for understanding, such as text annotations \citep{chowdhury_emergent_2021,martel_towards_2020,wang_tienet_2018,zhang_pathologist-level_2019} and intermediate feature visualizations \citep{bau_network_2017,natekar_demystifying_2020}, and more \citep{geirhos_imagenet-trained_2022,pisov_incorporating_2019,sun_saunet_2020}.
Notably, deriving attribution maps (or interpretation maps, saliency maps) is one of the most popular and meaningful manners for deep neural networks, and has attracted the interest of numerous works.

\begin{figure}[hbpt]
  \centering
  \includegraphics[width=.45\textwidth]{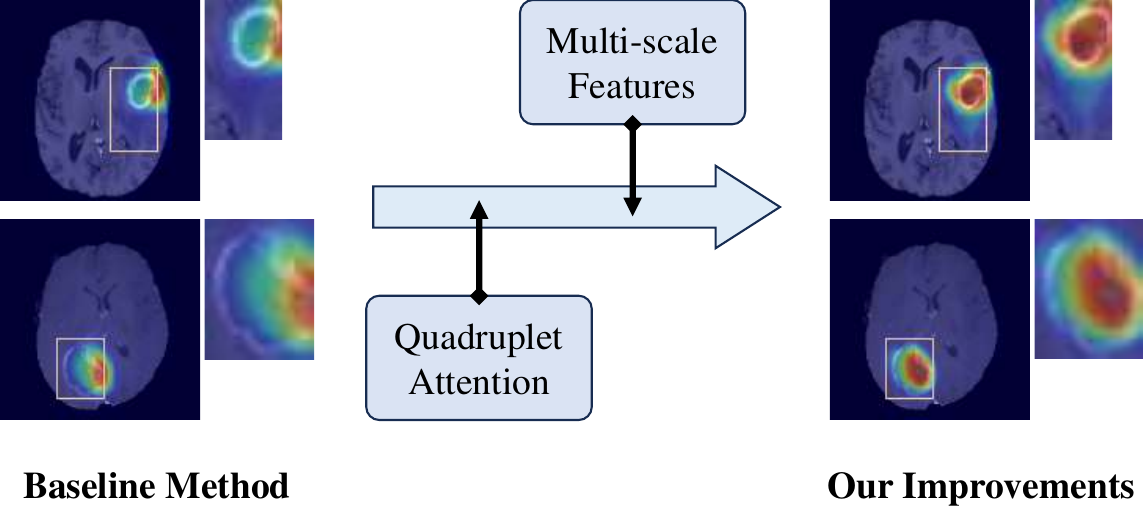}
  \caption{To enhance the localization capability of attribution maps, we propose to introduce quadruplet attention and mulit-scale features into prototypical part network.}
  \label{fig1:contributions}
\end{figure}

% post hoc method, concept-base, case-based 第一次解释
% vanilla 进一步解释，或者参考文献
% further attribution 的解释
% 网络名字没必要大写
A collection of presently popular techniques for attribution maps falls under the category of post hoc methods, such as class activation mapping (CAM) \citep{zhou_learning_2016} and its variants \citep{selvaraju_grad-cam_nodate,omeiza_smooth_2019}.
In post hoc methods, regardless of the network structure, the activated values within the attribution maps are obtained based on their contribution to the final outputs. Typically, the values are estimated using gradients or classification weights after training.
% CAM: Learning deep features for discriminative localization
% Grad-CAM: Grad-cam: Visual explanations from deep networks via gradient-based localization
% Grad-CAM++: Grad-CAM++: Improved Visual Explanations for Deep Convolutional Networks
% Smooth grad-cam++: Smooth grad-cam++: An enhanced inference level visualization technique for deep convolutional neural network models
These approaches enable the direct acquisition of attribution maps without incurring additional training costs or well-designed modifications to the architecture 
\footnote{CAM is an exception, since it requires a global average pooling layer at the end to calculate classification weights for the activated values}. 
Nevertheless, post hoc methods are commonly restricted to rough localization ability. 
In light of this, both concept-based and case-based methods attempt to establish explicit modules for achieving attribution maps. 
Concept-based methods, represented by concept bottleneck models (CBMs) \citep{koh_concept_2020}, constitute a series of annotation-based approaches. The concept-based methods typically necessitate a predefined dictionary containing intuitive and user-friendly concepts related to the classification task. Using the dictionary, decisions are usually made by evaluating the connections between the features and the given concepts, and the attribution maps are then represented based on these connections. Yet, the extra cost introduced by additional annotation is often unaffordable. 
On the other hand, case-based methods, represented by prototypical part networks (ProtoPNets) \citep{chen_this_2019}, is a category of self-explainable approaches. Akin to the concepts defined in the concept-based methods, case-based methods define a group of prototypes and calculate outputs and attribution maps based on the connections between features and prototypes. Instead, the prototypes are learned from the training data, thereby eliminating trivial yet costly handcrafted annotations.

% Part 3. CV 可解释性
% 将 general computer vision 和 medicine 的方法写成递进的关系
Many efforts in ProtoPNets of general computer vision are predominantly towards adjusting the decision process to closely resonate with human understanding. Illustratively, ProtoPShare \citep{rymarczyk_protopshare_2021} shares prototypes across classes while making decision; ProtoTree \citep{nauta_neural_2021} replaces the single-step decision based solely on the similarity with prototypes, by a decision path through the tree; ProtoPool \citep{rymarczyk_interpretable_2022} automatically assigns prototypes to different pools, refining prototype-based partially to pool-based decision; Deformable ProtoPNet \citep{donnelly_deformable_2022} adaptively changes the relative spatial positions depending on the input data, making it a deformable decision process. Additionally, some works also try to enhance the readability of the learned prototypes \citep{nauta_pip-net_2023}, or extend ProtoPNet into other tasks beyond classification \citep{sacha_protoseg_2023}.
% ProtoPShare: Prototype Sharing for Interpretable Image Classification and Similarity Discovery
% Neural Prototype Trees for Interpretable Fine-grained Image Recognition
% Pool: Interpretable Image Classification with Differentiable Prototypes Assignment
% Deformable ProtoPNet: An Interpretable Image Classifier Using Deformable Prototypes
% ProtoSeg: Interpretable Semantic Segmentation with Prototypical Parts
% PIP-Net: Patch-Based Intuitive Prototypes for Interpretable Image Classification
However, the increased focus on localization capability has also drawn considerable attention in medical imaging, where edge information holds paramount importance. Many works pay more attention on fine-grained attribution maps generated by ProtoPNets in medical imaging. 
XProtoNet \citep{kim_xprotonet_2021} employs a explicit pixel-level masking block as part of feature extraction module, forcing prototype vectors to capture fine-grained information. 
Building on this, MProtoNet \citep{wei_mprotonet_2024} introduces soft masking and online CAM loss into XProtoNet. Soft masking can sharpen and highlight the most prominent regions within the attribution maps, while online CAM aids in enhancing the localization capability of the pixel-level masking block during training.
Despite of their strengths, ongoing efforts to enhance localization performance is still a persistent focus.

% Part 3. 总结
% 3.1. 总结
% 贡献：将只能用在 2D 的拓展到 3D 解决了 XX 的问题
In this work, our principal goal is to enhance the localization capability within the existing prototype part networks for medical imaging. To this end, as depicted in Fig. \ref{fig1:contributions}, we introduce quadruplet attention layer, the extended version of triplet attention \citep{misra_rotate_2021} layer especially for 3D scenarios, as well as a multi-scale block into MProtoNet, naming the evolved network as \textbf{\underline{M}}ulti-scale \textbf{\underline{A}}ttentive \textbf{\underline{Proto}}typical Part \textbf{\underline{Net}}work (MAProtoNet). In our MAProtoNet, the novel designed quadruplet attention layers, whose 2D counterpart has been validated in many CAM-based post hoc interpretable methods, can improve the localization performance via augmenting the existing online CAM loss. Meanwhile, drawing inspiration from studies on image segmentation, where the introduction of multi-scale features has usually displayed outstanding improvement for pixel-level output, we propose a concise multi-scale block to generate attribution maps, forcing MAProtoNet to capture fine-grained features while also paying attention to coarse-grained yet significant information. Furthermore, we extend the original mapping loss proposed by XProtoNet to a multi-scale scenario, termed multi-scale mapping loss, to compel MAProtoNet to be robust to multi-scale affine transformation.
% Triplet
% Scalp-supervised contrastive learning for cardiopulmonary disease classification and localization in chest x-rays using patient metadata
% CrossEAI: Using Explainable AI to generate better bounding boxes for Chest X-ray images
% 3.2. 贡献
In brief, the contributions of our work can be summarized in the following three respects:
\begin{itemize}
  \item[1.] Firstly, we propose MAProtoNet, an interpretable model to enhance the localization capability for 3D magnetic resonance imaging (MRI) classification. Our MAProtoNet can achieve state-of-the-art localization performance in brain tumor classification.
  \item[2.] Secondly, we propose novel quadruplet attention layers and a multi-scale module, as well as the multi-scale mapping loss in our MAProtoNet. The quadruplet attention can provide attentive interactions for 3D MRI data, while the multi-scale module and the multi-scale mapping loss generate precise attribution maps through both fine-grained and coarse-grained information.
  \item[3.] Thirdly, we executed extensive experiments on brain tumor segmentation (BraTS) 2018, 2019, 2020 datasets, revealing that our proposed MAProtoNet exhibits considerable enhancement on localization as compared to its baseline methods. Our ablation experiments reveal insights into our multi-scale module. 
\end{itemize}

% Part 4. 结构
The remainder of this paper is structured as follows. Related works on multi-scales features and interpretable attribution methods for medical imaging are introduced in Section \ref{related works}. Definition of our problem and necessary preliminaries are given in Section \ref{preliminaries}. In Section \ref{proposed method}, our MAProtoNet is introduced. Experiments are given in \ref{experiments} and results are presented in Section \ref{results}. Our conclusions for this paper are in Section \ref{conclusion}.

\section{Related Works}
\label{related works}
% 重复不多，重新给缩写和参考文献
\subsection{Interpretable Attribution Methods}
Interpretable attribution methods, which aim to provide both classification results and the corresponding attribution maps for the decisions, can be primarily implemented through three kinds of approach in medical imaging: (a) post hoc methods; (b) concept-based methods and (c) case-based methods. 

\noindent
\textbf{(a) Post Hoc Methods.} Post hoc methods primarily derive attributions map by analyzing the contributions of input pixels or regions to the final decision. Many attempts to measure such contributions, especially the contribution scores for feature channels, have been widely explored. For example,  class activation mapping (CAM) \citep{zhou_learning_2016} explicitly designs a global average pooling layer so as to apply the weights of the final projection head as scores, while Grad-CAM \citep{selvaraju_grad-cam_nodate} employs the gradient directly for the weights. In most cases, the latter is more popular because it gets rid of meticulous designs which might usually reduce the performance of the original task.
Owing to its simplicity and intuitive nature, post hoc methods have been extensive applied in many scenarios in medical imaging, providing interpretability capabilities for works across tumor \citep{pereira_automatic_2018,frangi_weakly-supervised_2018}, schizophrenia \citep{lin_sspnet_2022}, and more \citep{kim_visual_2019,cancers13061291}. 
In particular, triplet attention \citep{misra_rotate_2021}, the technique which has been broadly employed in post hoc methods of 2D scenarios, have showed superior enhancement in localization. Existing works such as SCALP \citep{jaiswal_scalp_2021}, ChexRadiNet \citep{han_using_2021} and so forth \citep{zhao_crosseai_2023}, have proved the introduction of triplet attention can aid CAM and its variants in generating more precise attribution maps.

% Weakly-Supervised Learning-Based Feature Localization for Confocal Laser Endomicroscopy Glioma Images 
% Automatic Brain Tumor Grading from MRI Data Using Convolutional Neural Networks and Quality Assessment
% Visual Interpretation of Convolutional Neural Network Predictions in Classifying Medical Image Modalities: more problems
% Convolutional Neural Network-Based Clinical Predictors of Oral Dysplasia: Class Activation Map Analysis of Deep Learning Results
% SSPNet:An interpretable 3D-CNN for classification of schizophrenia using phase maps of resting-state complex-valued fMRI data

% SCALP-Supervised Contrastive Learning for Cardiopulmonary Disease Classification and Localization in Chest X-rays using Patient Metadata
% Using Radiomics as Prior Knowledge for Thorax Disease Classification and  Localization in Chest X-rays
% CrossEAI: Using Explainable AI to generate better bounding boxes for Chest X-ray images

\noindent
% 统一 many works
% this these afore
\textbf{(b) Concept-based Methods.}
Represented by concept bottleneck models (CBMs) \citep{koh_concept_2020}, concept-based methods break down the inference process into concept categorization and final classification. Benefiting from prior concepts usually annotated by human, concept-based methods can yield relatively better attribution maps in many cases \citep{patricio_coherent_2023,bie_mica_2024}. Yet, even with some attempts to leverage large language models (LLMs) for concept generation or reasoning \citep{yan_robust_2023,kim_concept_2023}, the costs associated with annotation, computation or knowledge barrier still pose challenges for the majority.

% Coherent Concept-based Explanations in Medical Image and Its Application to Skin Lesion Diagnosis
% MICA: Towards Explainable Skin Lesion Diagnosis via Multi-Level Image-Concept Alignment

% Robust and Interpretable Medical Image Classifiers via Concept Bottleneck Models: GPT-based concepts
% Concept Bottleneck with Visual Concept Filtering for Explainable Medical Image Classification: LLM

\noindent
% double check
\textbf{(c) Case-based Methods.}
Prototypes, similar to concepts in concept-based methods, are applied to establish connections between features and human understandings in case-based methods. However, prototypes in self-explained case-based methods are typically learned from training data rather than predefined. A number of works have introduced case-based methods directly in medical imaging problems \citep{mohammadjafari_using_2021,rousseau_applicability_2023}. Additionally, to refine the structure for achieving more precise maps, XProtoNet \citep{kim_xprotonet_2021} introduces a mapping module for pixel-level mask generation into ProtoPNet for chest radiography diagnosis. A mapping loss is also applied to encourage attribution maps to be robust against affine transformations. MProtoNet \citep{wei_mprotonet_2024} steps further to achieve interpretable 3D brain tumor classification. The online CAM loss is only employed during training, to supervise the mask generation module for precise localization. In addition, the out-of-the-box soft masking technique is also introduced in MProtoNet to sharpen and highlight the values of the attention maps. 

In this work, inspired by triplet attention in post hoc methods, we extend it into quadruplet attention for our 3D scenarios, enhancing the localization capability via existing online CAM loss. 

\subsection{Multi-scale Features for Medical Image Segmentation}
Image segmention can be formulated as the problem of classifying pixels with pixel-level labels (semantic, instance or panoptic segmentation) \citep{minaee_image_2021}. Since the ability to capture long-range pixel dependency, many researches have demonstrated the significant benefits of multi-scale features for such pixel-level tasks.
% Image Segmentation Using Deep Learning: A Survey  (definition)
One of the most renowned categories of multi-scale methods in image segmentation is exemplified by UNet \citep{ronneberger_u-net_2015} and its wide-ranging variants \citep{li_h-denseunet_2018,huang_unet_2020}. In their encoder-decoder design, coarse-scale features are integrated into fine-scale features directly by concatenating along the channels. As transformers gain popularity, novel multi-scale architectures like Mask2Former \citep{cheng_masked-attention_2022,cheng_per-pixel_2021} also unveil its impact in recent years. In their architecture, leveraging single-scale or multi-scale features from traditional CNNs, an additional transformer decoder module generates pixel-wise masks using multiple cross-attention layers. Moreover, to cope with the computational costs stemming from multi-scale features, deformable attention \citep{zhu_deformable_2021} is usually introduced in these works. 
% MP-Former
% deformable detr
Even within the field of medical imaging, these transformer-based multi-scales models persist as a common practice for segmentation \citep{yuan_devil_2023}.
% Devil is in the Queries: Advancing Mask Transformers for Real-world Medical Image Segmentation and Out-of-Distribution Localization

Actually, multi-scale features have also been utilized in ProtoPNets in computer vision (without a pixel-level mapping module), to extract multi-level prototypes, proving that fine-grained information is useful in this framework \citep{wang_hqprotopnet_2023}. Consequently, in this work, we devise a concise block to take advantage of multi-scale features from CNNs for generating pixel-level masks in our MAProtoNet.
% HQProtoPNet: An Evidence-Based Model for Interpretable Image Recognition

\section{Preliminary}
\label{preliminaries}
\begin{figure*}[ht]
  \centering
  % backbone, as feature extraction
  \includegraphics[width=.9\textwidth]{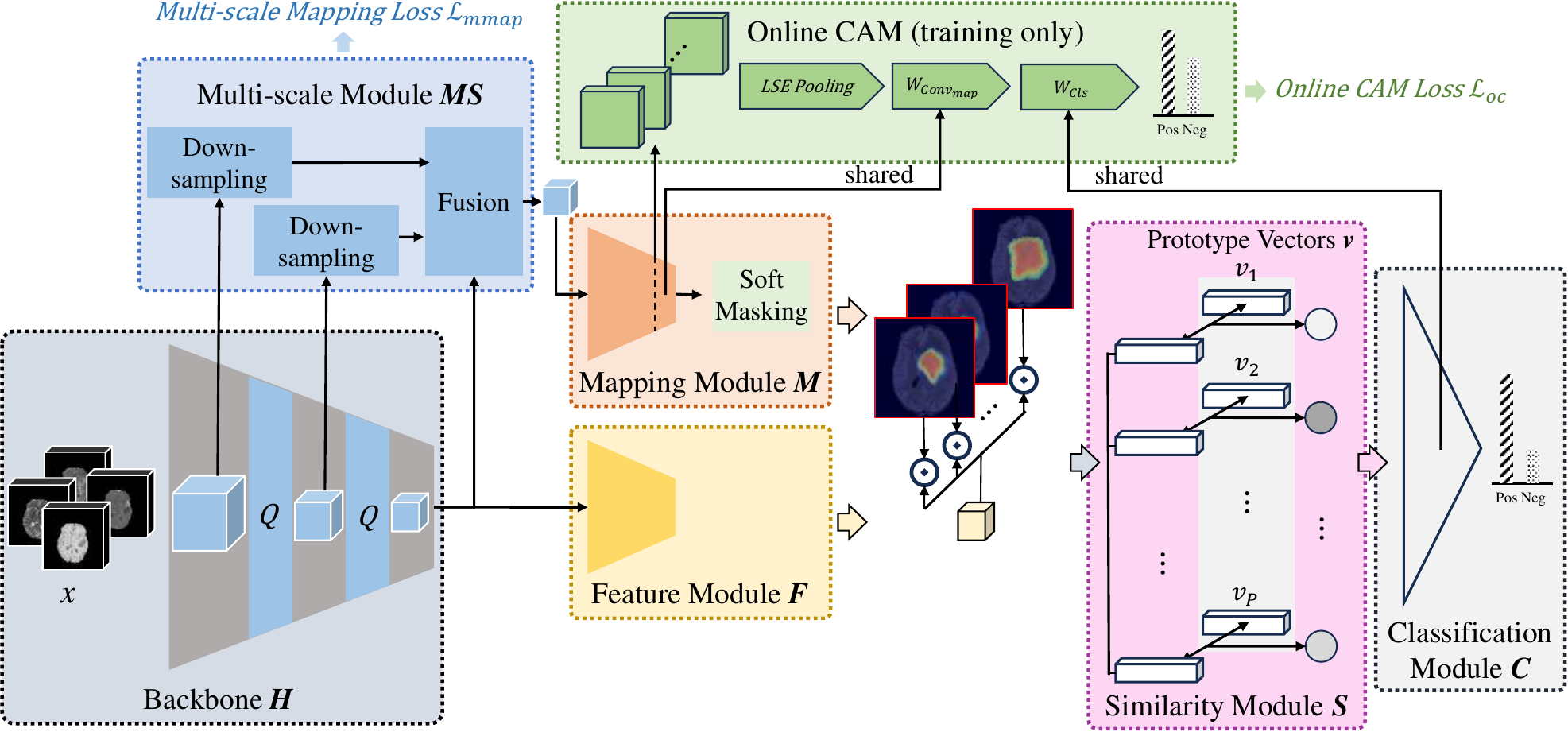}
  \caption{Framework of our proposed MAProtoNet. The backbone $H$ is mainly for feature extraction. Green emphasizes the enhancements of XProtoNet by \textcolor{green}{MProtoNet}, whereas blue is used to highlight further improvements by our \textcolor{blue}{MAProtoNet}. In our architecture, quadruplet attention blocks denoted by $Q$ are introduced, and multi-scale features are applied and fused in a novel multi-scale module for pixel-level maps generation. We would like to emphasize that while we illustrate $n_{scale}=3$ in the figure, we actually set $n_{scale}=2$ in practice due to the shallower backbone.}
  \label{fig2:framework}
\end{figure*}
In this section, we formulate the interpretable attribution problem from the perspective of prototypical part networks in medical imaging, and the original triplet attention operation as well.
\subsection{Prototypical Part Networks in Medical Imaging}
% In this section 引子
% 不用将来时
% 公式作为句子成分
% 统一 medicine 和 medical image
Given a MRI dataset 
$T=\{(x_1, y_1), (x_2, y_2), ..., (x_n, y_n)\} \in (\mathcal{X}, \mathcal{Y}) ^ n$, 
$x \in \mathcal{X}$ 
denotes the pre-processed input, while 
$y \in \mathcal{Y}$ represents the classification labels in space $\{0, 1, ..., C\}$. 
With a prototypical part network $PrototypicalPartNet$ and the number of prototypes denoted as $P$, we have:
\begin{align}
    y^{pred}, map = PrototypicalPartNet(x),
    \label{eq1}
\end{align}
where $y^{pred}$ denotes the output logits, and $map=(map^1, map^2, ..., map^P)$ is a collection of $P$ resulting attribution maps, facilitating further calculation of similarity scores with $P$ learnable prototype vectors.

% . Here ... 或者 , where
% L1 norm

Specifically, as shown in Fig. \ref{fig2:framework} but without quadruplet attention layers and multi-scale module, $PrototypicalPartNet$ in medicine imaging is usually composed of four learnable modules: $H(x), F(x), M(x)$ and $C(x)$, standing for the sub-networks for backbone, deeply features extraction, pixel-level maps generation and classification, respectively. Hence, while introducing another unlearnable similarity function $S(x,y)$, as well as the learnable prototype vectors $v=(v_1, v_2, ..., v_P)$, Eq. \ref{eq1} can be further elucidated as Eq. \ref{eq2} below.
\begin{subequations}\label{eq2}
\begin{align}
    &\text{$h=H(x)$} \label{eq2:bb}, \\
    &\text{$\underline{map}=M(h)$} \label{eq2:attr}, \\
    &\text{$fea=F(h)$} \label{eq2:fea}, \\
    &\text{$scores=S(map \odot fea, v)$}, \label{eq2:sim} \\
    &\text{$\underline{y^{pred}}=C(scores)$}. \label{eq2:head}
\end{align}
\end{subequations}
Here $\odot$ is element-wise product operation.

% lambda 列出，不用 lambdas
% Lmap 自监督 -> Lmap 不用使用 map
In conjunction with regular cross-entropy loss, as well as the loss for learnable prototypes, mapping loss proposed in XProtoNet presented in Eq. \ref{eq3} and online CAM loss proposed in MProtoNet contribute to the more robust and precise attribution maps. 
\begin{align}
    \mathcal{L}_{map} = 
    \sum_{p \in P}
    {
    \left \|   
     M(A(h))_{p} - A(M(h)_{p})
    \right \|_1
    }.
    \label{eq3}
\end{align}
Here $A$ denotes the affine transformation, and $\left \| \cdot \right \|_1$ denotes the L1 norm. Hence, the total loss will be formulated as Eq. \ref{eq4}.
\begin{align}
    \mathcal{L} = 
    \underbrace{\mathcal{L}_{cls}}_{classification} + 
    \underbrace{\lambda_{clst} \mathcal{L}_{clst} + \lambda_{sep} \mathcal{L}_{sep}}_{protorypes} + 
    \underbrace{\lambda_{map} \mathcal{L}_{map}}_{robustness} + 
    \underbrace{\lambda_{oc} \mathcal{L}_{oc}}_{localization},
    \label{eq4}
\end{align}
where $\mathcal{L}_{cls}$ is the cross-entropy loss on the classification label $y$, $\mathcal{L}_{clst}$ and $\mathcal{L}_{sep}$ are cluster and separation losses for learnable prototype vectors $v$, $\mathcal{L}_{oc}$ is the online CAM loss on label $y$, $\lambda_{clst}, \lambda_{sep}, \lambda_{map}$ and $\lambda_{oc}$ are their corresponding weights. Since $\mathcal{L}_{map}$ defined in Eq. \ref{eq3} is also independent of the ground truth of segmentation, annotations related to segmentation maps are left out during the whole training process.
% In this work, $x_i\in \mathbb{R} ^{M \times H \times W \times D}, i=\{1, 2, ..., n\} $ is a pre-processed multi-parametric 3D magnetic resonance imaging, where $M$ denotes the number of modalities here, while $H$, $W$, $D$ are the spatial dimensions for 3D medical imaging data.
% four modalities: T1-weighted, T1-weighted contrast enhancement (T1CE), T2-weighted and T2 fluid attenuated inversion recovery (FLAIR).
% SE 和 CBAM 有没有必要突出
\subsection{Triplet Attention}
Triplet attention is the strategy to build connections across dimensions. Distinguished from the channel-wise squeeze-and-excitation networks \citep{wang_hqprotopnet_2023}, and convolutional block attention module \citep{hu_squeeze-and-excitation_2019} that calculates channel-wise and spatial-wise attention separately, triplet attention alternates attention across channel, height, width, denoted as $(C, H, W)$, via capturing dependencies between pairs-wise attention of $(H, W)$, $(C, W)$ and $(C, H)$, respectively. The resulting output is the average on three items, so as to attend all possible interactions of all three dimensions. 
% Rotate to Attend: Convolutional Triplet Attention Module

% take -> suppose
% 2D image with + channel
Mathematically, suppose we take spatial-wise attention or interaction cross $(H, W)$ pair as an illustration, given a 2D natural image $x \in \mathbb{R} ^ {C \times H \times W}$ with the channel dimension $C$ placed in the first order, the Z-pool function is defined in Eq. \ref{eq5}:
\begin{align}
    Z \text{\textendash} pool(x) = \left [ MaxPool_{1d}(x), AvgPool_{1d}(x) \right ],
    \label{eq5}
\end{align}
% , 写 concat。0d -> first dimension
where $MaxPool_{1d}$ and $AvgPool_{1d}$ represent the pooling operations occurring across the first dimension ($C$ in this example).
The formulation of the spatial-wise branch in triplet attention is as Eq. \ref{eq6}:
% x 和函数之间加个 点
\begin{align}
    Tri_{HW}(x) = x \cdot \psi (Z \text{\textendash} pool(x)).
    \label{eq6}
\end{align}
$\psi$ represents the standard 2D convolutional layers to focus on interactive information within the remaining dimensions ($H$ and $W$ in this example). Finally, the output of triplet attention block is outlined by Eq. \ref{eq7}:
\begin{align}
    Tri(x)= \frac{1}{3}  (Tri_{HW}(x) + Tri_{CW}(x^{H}) + Tri_{CH}(x^{W})).
    \label{eq7}
\end{align}
$x^{H}$ and $x^{W}$ are reshaped from original input $x$, with the $H$ and $W$ dimensions placed in the first order, respectively. Here, we also assume that the outputs of $Tri_{CW}$ and $Tri_{CH}$ are restored back to the original shape.

\section{The Proposed Method}
\label{proposed method}
Framework of our MAProtoNet is given in Fig. \ref{fig2:framework}. Stemming from MProtoNet \citep{wei_mprotonet_2024}, our MAProtoNet incorporates additional quadruplet attention layers and a multi-scale module, and a novel multi-scale mapping loss as well. Thus, with respect to our improvements, we organize this section into three parts: quadruplet attention, multi-scale module and multi-scale mapping loss.

\subsection{Quadruplet Attention}
\begin{figure}[ht]
  \centering
  \includegraphics[width=0.5\textwidth]{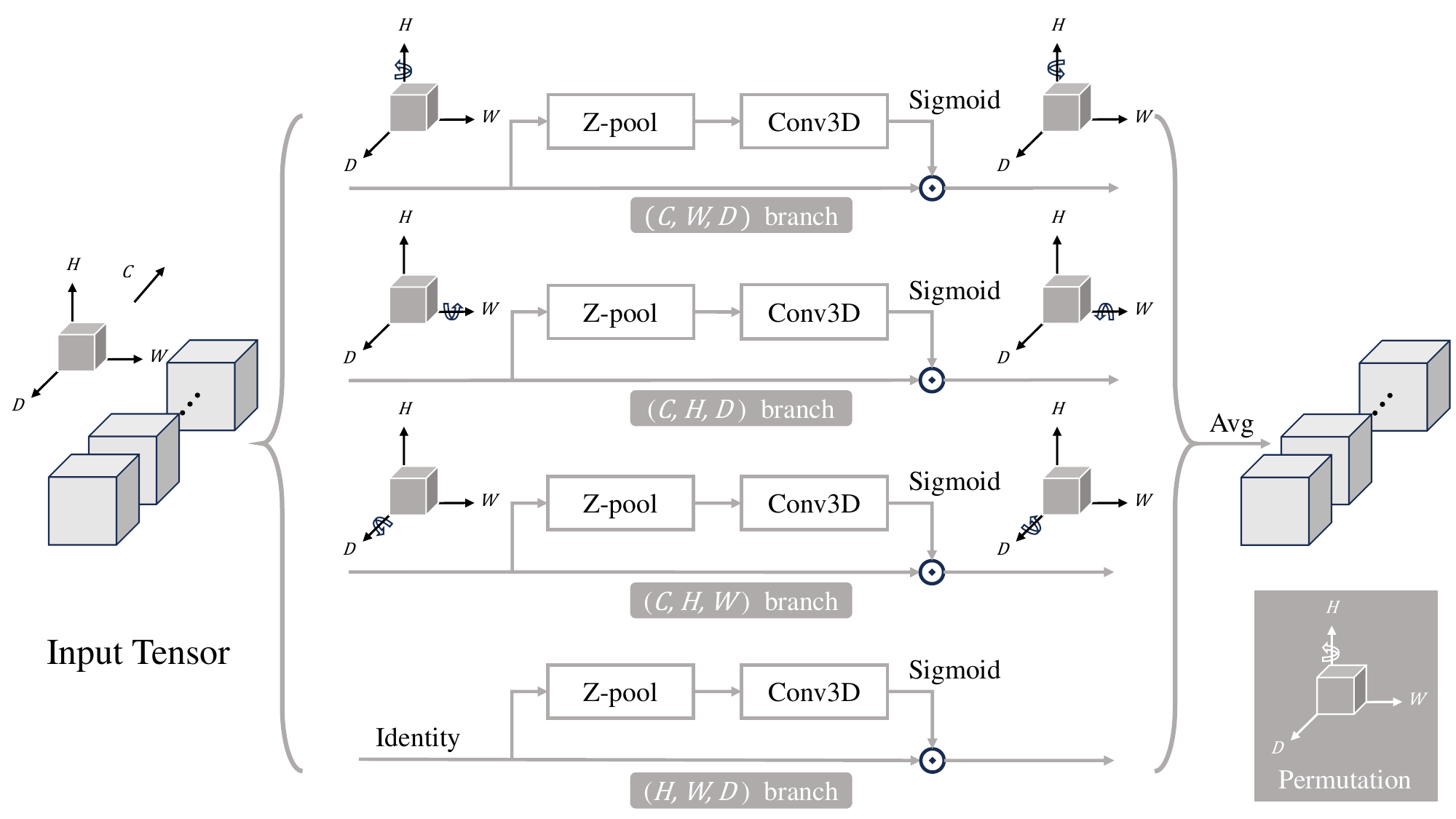}
  \caption{Illustration of the proposed quadruplet attention. Distinct interactions are extracting via four different branches, and are fused by averaging.}
  \label{fig3:quadruplet}
\end{figure}

In order to introduce 2D triplet attention for boosting the online CAM loss for our 3D medical imaging data, a natural thinking is to extend its triplet operations to quadruplet operations. To this end, we draw inspiration from triplet attention, to rotate and attend interactions cross $(C, H, W, D)$ dimensions, D here denotes the depth of 3D data. 
Unlike vanilla triplet attention block where pairs are applied to extract dependencies, our quadruplet attention block is named so because it employs triplets from $(H, W, D), (C, W, D), (C, H, D)$ and $(C, H, W)$ to capture interactions through 3D convolutional layers, yielding the average output over four items. 
As illustrated in Fig. \ref{fig3:quadruplet}, the quadruplet attention consists of four branches, each capturing diverse interactions via triplets across different dimension combination. In the first three branches, rotation operation is adopted at the beginning and the restoration operation follows at the end, so as to establish connections between $C$ and other dimensions.

Mathematically, similar to triplet attention, given a 3D medical imaging data $X \in \mathbb{R}^{C \times H \times W \times D}$ and the Z-pool function as in Eq. \ref{eq5} as well, the spatial-wise branch in Eq. \ref{eq6} is updated as:
\begin{align}
    Quad_{HWD}(X) = X \cdot \Psi (Z \text{\textendash} pool(X)),
    \label{eq8}
\end{align}
where $\Psi$ represents the standard 3D convolutional layers concentrating on the remaining dimensions after Z-pool. We can then derive Eq. \ref{eq9} as:
\begin{equation}
\begin{aligned}
    Quad(X)= & \frac{1}{4}  (
    Quad_{HWD}(X) 
    + Quad_{CWD}(X^{H}) \\ & 
    + Quad_{CHD}(X^{W})  
    + Quad_{CHW}(X^{D})
    ).
    \label{eq9}
\end{aligned}
\end{equation}
Similarly, $X^{H}$, $X^{W}$ and $X^{D}$ represent the permuted original inputs, with $H$, $W$ and $D$ placed in the first order, respectively. Permutation operations are also positioned at the end of $Quad_{CWD}$, $Quad_{CHD}$ and $Quad_{CHW}$ function to restore the original shape. 
\subsection{Multi-scale Module} 
\begin{figure*}[ht]
  \centering
  \includegraphics[width=0.9\textwidth]{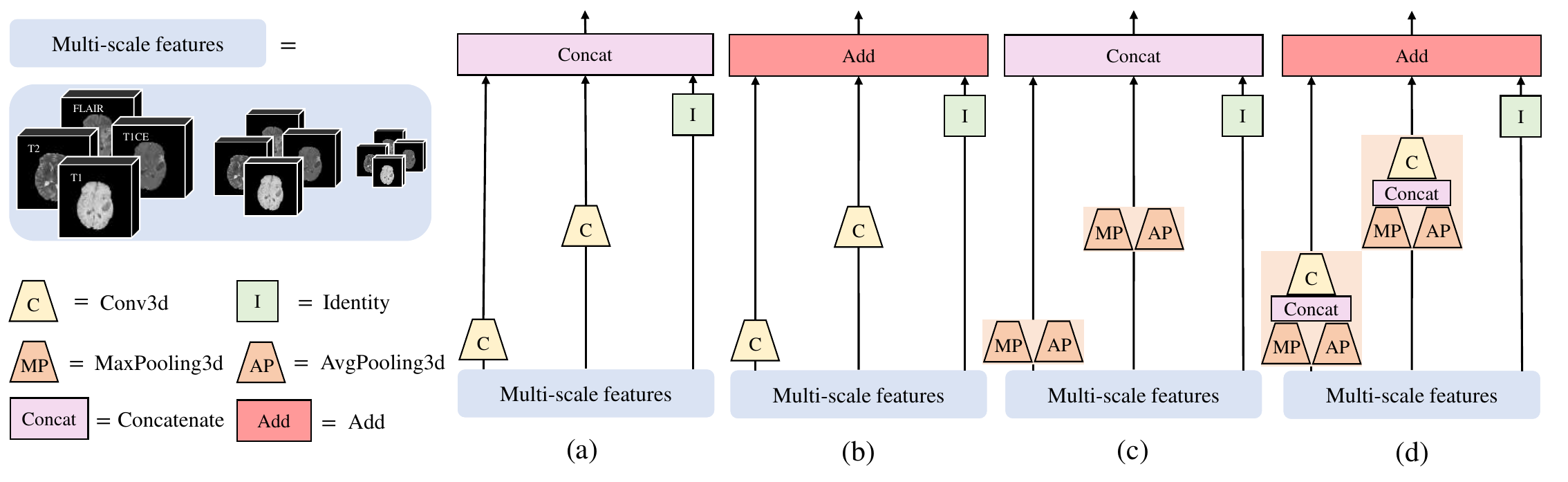}
  \caption{Architecture of the multi-scale module. As for the coarse-grained features, we down-sample via convolutional or pooling layers, and fuse them by an addition or concatenation operation. Hence, we have architecture (a) $Conv + Concat$; (b) $Conv + Add$; (c) $Pool + Concat$ and (d) $Pool + Add$. Convolutional layers in (a) and (b) are designed to decrease spatial resolution with a $stride$ greater than 1, whereas those in (d) are channel-wise with $kernel\_size=1$, specifically to maintain consistency in the channel dimension.}
  \label{fig4:multi-scale}
\end{figure*}

% MS 的顺序
\label{multiscales}
In order to provide both fine-grained and coarse-grained information simultaneously for generation of attribution maps, we directly reuse the attentive intermediate outputs from backbone as the multi-scale features, obtaining a more comprehensive overall visual representation. To utilize these features effectively, as depicted in Fig. \ref{fig4:multi-scale}, we design architectures for coarse-grained feature extraction and fusion. Particularly, we down-sample coarse-grained features via convolutional or pooling layers, and then fuse them with the fine-grained information through addition or concatenation operation, resulting in four possible variants represented as (a), (b), (c) and (d) in Fig. \ref{fig4:multi-scale}. Subsequently, tensors containing both fine-grained and coarse-grained information are forwarded to the map module for further integration and attribution maps generation. Hence, with the introduced multi-scale module $MS(x)$, Eq \ref{eq2:bb} $\sim $ \ref{eq2:fea} is updated by:
\begin{subequations}\label{eq10}
\begin{align}
    &\text{$h^{1}, ..., h^{S}=H(x)$} \label{eq10:bb}, \\
    &\text{$h^{mul}=MS(h^{1}, ..., h^{S})$} \label{eq10:mul}, \\
    &\text{$map=M(h^{mul})$} \label{eq10:attr}, \\
    &\text{$fea=F(h^{S})$} \label{eq10:fea}.
\end{align}
\end{subequations}
Here we suppose the number of feature scales $n_{scale}=S$. It is worth noting that the fine-grained features $h^{S}$ are also directly concatenated or added onto the augmented tensors, while the following module learns to select information of interest for further attention. In essence, the original single-scale features are a specific case within our augmented features, with corresponding coarse-grained information or their weights being 0.

\subsection{Multi-scale Mapping Loss}
In our MAProtoNet, we extend the mapping loss given in Eq. \ref{eq3}, to the multi-scale mapping loss $\mathcal{L}_{mmap}$ formulated in Eq. \ref{eq11}. As introduced in Section \ref{multiscales}, we employed the multi-scale features in our MAProtoNet. Accordingly, to endow the model with robustness to affine transformations, we intuitively generalize the original mapping loss, which only considers the robustness against perturbations to the fine-grained features $h^{S}$, to encompass robustness to features at all scales $(h^{1}, ..., h^{S})$. 
Naturally, the multi-scale fusion output is calculated by the transformed features as in Eq. \ref{eq11:mulmap}, and the multi-scale mapping loss $\mathcal{L}_{mmap}$ as depicted in Eq. \ref{eq11:mulloss} comes from the resulting fusion output $^{A}h^{mul}$.
\begin{subequations}\label{eq11}
\begin{align}
    &\text{$^{A}h^{mul}=MS(A(h^{1}), ..., A(h^{S}))$} \label{eq11:mulmap}, \\
    &\text{$
    \mathcal{L}_{mmap}=
    \sum_{p \in P}
    \left \|
    M(^{A}h^{mul})_{p} 
     - A(M(h^{mul})_{p})
    \right \|_1
    $}.
    \label{eq11:mulloss}
\end{align}
\end{subequations}
% \begin{equation}
% \begin{aligned}
%     & \mathcal{L}_{mmap} = \\
%     & \sum_{p \in P}
%     \left \|
%     % \begin{aligned}
%     f_{attr}(
%             A(^{1}x^{fea}_i), ..., A(^{S}x^{fea}_i)
%         )_{p} 
%     - A(
%         f_{attr}(h_i^{mul})_{p}
%     ) 
%     % \end{aligned}
%     \right \|_1
%     \label{eq11}
% \end{aligned}
% \end{equation}
% 解释 A
Here we denote the multi-scale fusion output conditioned on the transformed multi-scale features as $^{A}h^{mul}$, $A$ is the affine transformation similar to the definition in Eq. \ref{eq3}. Eventually, the total loss of our model is developed into Eq. \ref{eq12} as below.
\begin{align}
    \mathcal{L} = \mathcal{L}_{cls} + \lambda_{clst} \mathcal{L}_{clst} + \lambda_{sep} \mathcal{L}_{sep} + \underbrace{\lambda_{mmap} \mathcal{L}_{mmap}}_{ours} + \lambda_{oc} \mathcal{L}_{oc},
    \label{eq12}
\end{align}
where $\lambda_{mmap}$ is the weight for our multi-scale mapping loss.

\noindent
\textbf{Summarization.} As in Fig. \ref{fig2:framework}, compared to existing prototypical part networks for medical imaging such as XProtoNet \citep{kim_xprotonet_2021} and MProtoNet \citep{wei_mprotonet_2024}, our MAProtoNet shares similar pipeline for classification, comprising a backbone, a mapping module, a feature module, a similarity module and a classification module. Similarly to MProtoNet, our MAProtoNet also integrates online CAM and soft masking into its architecture. On the flip side, we introduce quadruplet attention layers $Q$ into the backbone to capture attentive multi-scale features, and propose a multi-scale module $MS$ for multi-scale features down-sampling and fusion. The fusion output, combining both fine-grained and coarse-grained information, is utilized for map generation instead of relying solely on the fine-grained feature. A novel multi-scale mapping $\mathcal{L}_{mmap}$ is also proposed to supervise the multi-scale module.

\section{Experiments}
\label{experiments}
\subsection{Datasets}
Brain tumor segmentation challenge is a famous benchmark annually published by medical image computing and computer assisted intervention (MICCAI) community since 2012. BraTS has emerged as one of the foremost benchmarks to evaluate state-of-the-art methods for brain tumor segmentation. In this work, we adopt BraTS 2018 \footnote{https://www.med.upenn.edu/sbia/brats2018/}, 2019 \footnote{https://www.med.upenn.edu/cbica/brats-2019/} and 2020 \footnote{https://www.med.upenn.edu/cbica/brats2020/} datasets in experiments. Each includes several clinically-acquired 3T multi-parametric magnetic resonance imaging (mpMRI) scans, comprising four 3D MRI modalities: T1, T1CE, T2 and FLAIR. These scans are rigidly aligned, resampled to $1 \times 1 \times 1$ mm$^{3}$ isotropic resolution, skull-stripped, with final dimensions of $240 \times 240 \times 155$. It should be emphasized that despite discrepancies among these three datasets, data overlap occurs as each iteration is built upon its predecessor. Nonetheless, these datasets remain valuable benchmarks for studying generalization. Details of the datasets we use are shown in Tab \ref{tab1}. Due to the absence of segmentation annotations, which could be used exclusively for validation purposes, we only employ the training set in experiments.

% previous
\begin{table}[htbp]
  \centering
  \caption{Details of BraTS datasets. HGG, LGG denote high-grade glioma and low-grade glioma, respectively. Num. of Overlap indicates the number of subjects that overlap with their previous predecessors.}
  \begin{tabular}{ccccc}
    \toprule
    \multirow{2}{*}{Dataset} & \multicolumn{3}{||c||}{Num. of Subjects} & \multirow{2}{*}{Num. of Overlap} \\
     & \multicolumn{1}{||c}{Train} & \multicolumn{1}{|c}{HGG} & \multicolumn{1}{c||}{LGG} & \\
    \midrule
    BraTS 2018 & \multicolumn{1}{||c}{285} & \multicolumn{1}{|c}{210} & \multicolumn{1}{c||}{75} &  N/A \\
    BraTS 2019 & \multicolumn{1}{||c}{335} & \multicolumn{1}{|c}{259} & \multicolumn{1}{c||}{76} & 285 (85.07\%) \\
    BraTS 2020 & \multicolumn{1}{||c}{369} & \multicolumn{1}{|c}{293} & \multicolumn{1}{c||}{76} & 167 (45.25\%) \\
    \bottomrule
  \end{tabular}
  \label{tab1}
\end{table}

Within the datasets, the MRI voxels of each subject from diverse institutions and scanners is marked by 3 regions: whole tumor (WT), tumor core (TC) and enhancing tumor (ET). As similar to the configuration in MProtoNet, we apply WT regions as ground truth for evaluation of localization ability. We also follow their configurations for data pre-processing. We crop the raw MRI images into voxels shaped $192 \times 192 \times 144$, then down-sample them to $128 \times 128 \times 96$, and pad all 4 modalities into channel dimension, resulting in $x \in \mathbb{R}^{4 \times 128 \times 128 \times 96}$ as input. Details for further data pre-processing and augmentation can be found in \ref{app:data}.
\subsection{Baselines}
Since our work is rooted in MProtoNet, we follow their experiment settings, and implement a typical 3D CNN with residual connections \citep{he_deep_2016}, ProtoPNet \cite{chen_this_2019}, XProtoNet \citep{kim_xprotonet_2021}, as well as MProtoNet \citep{wei_mprotonet_2024} itself, as our baseline methods. All four baselines can be regarded as a submodule of our MAProtoNet in Fig. \ref{fig2:framework}. Attribution maps of CNN are obtained via Grad-CAM, while ProtoPNet are interpreted by its prototypes. XProtoNet, MProtoNet and our MAProtoNet generate maps from the mapping module directly. 

% 不加具体，直接引用
% \noindent
% \textbf{CNN.} CNN baseline is a submodule only with a backbone and a feature module. The resulting features are pooled by a global average pooling layer and subsequently classified using a linear head. Attribution maps are calculated from the final convolutional layer via Grad-CAM.

% \noindent
% \textbf{\textbf{ProtoPNet.}} ProtoPNet is the network without mapping module for pixel-level mask generation. The attribution maps are defined by similarity scores between the features and the learned prototype vectors.

% \noindent
% \textbf{XProtoNet.} Structure of XProtoNet is the similar to MAProtoNet, without the training strategies depicted in green and blue in Fig. \ref{fig2:framework}. Online CAM, soft masking, quadruplet attention and multi-scale module are absent. 

% \noindent
% \textbf{MProtoNet.} As compared with our MAProtoNet, MProtoNet fails to capture attentive multi-scale features for mapping module in a multi-scale module.

% previous
\subsection{Evaluation metrics}
We analyse our MAProtoNet from three aspects: classification, localization as well as interpretability, as similar to our previous work. Considering the imbalance between positive and negative samples in the datasets, we employed balanced accuracy (BAC) to evaluate the correctness. We calculate the activation precision (AP) to measure the localization coherence. We also apply incremental deletion score (IDS) as an additional metric to estimate the interpretability performance. Details of calculation of these metrics can be found in \ref{app:metrics}.
% IAIA-BL: A Case-based Interpretable Deep Learning Model for Classification of Mass Lesions in Digital Mammography.
\subsection{Implementation}
In the experiments, we introduce 3D ResNet152 \citep{he_deep_2016} for feature extraction. In our MAProtoNet, we only use the input layers, as well as the first residule block (four blocks in total in ResNet152) as our backbone. For all prototypical part networks, we make use of 30 prototypes, with 15 prototypes assigned to each category. Dimension of each prototype vector is set to 128. The coefficients of each loss item defined in Eq. \ref{eq4} (if has) are: $\lambda_{cls}=1, \lambda_{clst}=0.8, \lambda_{sep}=0.08, \lambda_{map}=0.5, \lambda_{OC}=0.05, \lambda_{L1}=0.5$. We replace the original mapping loss with multi-scale mapping loss as formulated in Eq. \ref{eq12} in our MAProtoNet, with $\lambda_{mmap}=0.5$ too. The number of scales we use is $n_{scales}=2$. All of these parameters are configured based on the experiences and settings in MProtoNet.

We also follow latest practices, using AdamW optimizer with a baseline learning rate of 0.001 and a weight decay of 0.01. We train 100 epochs in total with a batch size of 32. Learning rate scheduler is utilized with linear warm-up for the first 20 epochs and cosine annealing for the remaining 80 epochs. In the training process, we implement an extra pushing operation after every 10 epochs, and additional 10 epochs solely for linear head on cross-entropy and L1 loss after pushing. More details for the training process are provided in \ref{app:training}. We conducted our experiments on a NVIDIA RTX A6000 GPU, with each configuration running for approximately 6$\sim$8 hours (5-fold cross validation).

% ResNet 参考
% first two blocks
% shallower 清晰
% lambda 基于 experience 。。。
% 解释 push 或者给 Appendix

\noindent 
\textbf{Discussion.} We want to highlight that compared to the backbone reported in \citep{wei_mprotonet_2024} which employs the first two residual blocks, we only use the first one block for features extraction. This configuration arises from our experimental findings indicating the occurrence of overfitting. We also notice a similar occurrence of overfitting in MProtoNet. We conjecture that the integration of online CAM, along with the multi-scale module, may have contributed to this outcome. Thus, for MProtoNet and our MAProtoNet, we only use the first residual block as backbone. For other baseline models, we keep the same configurations as in \citep{wei_mprotonet_2024}. Additionally, we also observe that ProtoPNet achieve significantly diminished performance in both localization and interpretability as reported by MAProtoNet, since they seem to calculate the attribution maps with inverted values in their code. Consequently, we fit them and report the correct results. We also present their original reported results in our results.

\section{Results and Analysis}
\label{results}
\begin{table*}[htbp]
  \centering
  \caption{Results from three perspectives: correctness, localization and interpretability. Results of our MAProtoNet here are achieved by architecture Fig. \ref{fig4:multi-scale}(c). In the table, $Q$ and $MS$ denote our proposed quadruplet attention and multi-scale module, respectively. Small font values in BraTS 2020 dataset are results reported by MProtoNet. Text in red represents \textcolor{red}{improvement} compared with MProtoNet, whereas green denotes \textcolor{green}{decreasement}.}
  \begin{tabular}{c||c||ccc}
    \toprule
    \multirow{2}{*}{Dataset} & \multirow{2}{*}{Method} & \multicolumn{1}{c}{$_{Correctness}$} & \multicolumn{1}{c}{$_{Localization \& Interpretability}$} & \multicolumn{1}{c}{$_{Interpretability}$} \\
    & & BAC (\%) $\uparrow$ & AP (\%) $\uparrow$ & IDS (\%) $\downarrow$ \\
    % 2018
    \midrule
    \multirow{7}{*}{BraTS 2018} & CNN & $85.5 \pm 2.2$ & $7.4 \pm 2.4$ & $11.3 \pm 3.5$ \\
    & ProtoPNet & $86.2 \pm 5.3$ & $13.7 \pm 5.6$ & $23.9 \pm 12.4$ \\
    & XProtoNet & $87.6 \pm 3.5$ & $15.3 \pm 1.6$ & $15.9 \pm 4.1$ \\
    & MProtoNet & $86.8 \pm 4.3$ & $80.9 \pm 3.2$ & $4.6 \pm 3.2$ \\
    & MProtoNet + $Q$ (ours) & ${86.6 \hspace{2pt} (\textcolor{green}{-0.2}) \pm 4.7}$ & ${81.2 \hspace{2pt} (\textcolor{red}{+0.3}) \pm 3.1}$ & ${6.3 \hspace{2pt} (\textcolor{green}{+1.7}) \pm 3.3}$ \\
    & MProtoNet + $MS$ (ours) & ${87.0 \hspace{2pt} (\textcolor{red}{+0.2}) \pm 2.9}$ & ${79.0 \hspace{2pt} (\textcolor{green}{-1.9}) \pm 4.0}$ & ${4.5 \hspace{2pt} (\textcolor{red}{+0.1}) \pm 2.5}$ \\
    & MAProtoNet (ours) & ${87.7 \hspace{2pt} (\textcolor{red}{+0.9}) \pm 3.7}$ & ${\textbf{82.2} \hspace{2pt} (\textcolor{red}{+1.3}) \pm 6.2}$ & ${\textbf{4.1} \hspace{2pt} (\textcolor{red}{-0.5}) \pm 2.3}$ \\
    % 2019
    \midrule
    \multirow{7}{*}{BraTS 2019} & CNN & $88.7 \pm 2.8$ & $12.6 \pm 3.3$ & $8.3 \pm 3.2$ \\
    & ProtoPNet & $88.0 \pm 4.2$ & $16.0 \pm 1.9$ & $22.5 \pm 8.4$ \\
    & XProtoNet & $\textbf{90.0} \pm 2.6$ & $18.8 \pm 2.6$ & $28.2 \pm 14.6$ \\
    & MProtoNet & $88.2 \pm 4.5$ & $81.1 \pm 2.2$ & $8.2 \pm 8.1$ \\
    & MProtoNet + $Q$ (ours) & ${88.0 \hspace{2pt} (\textcolor{green}{-0.2}) \pm 4.5}$ & ${82.5 \hspace{2pt} (\textcolor{red}{+1.4}) \pm 5.1}$ & ${\textbf{6.1} \hspace{2pt} (\textcolor{red}{-2.1}) \pm 4.3}$ \\
    & MProtoNet + $MS$ (ours) & ${87.5 \hspace{2pt} (\textcolor{green}{-0.7}) \pm 2.8}$ & ${79.0 \hspace{2pt} (\textcolor{green}{-2.1}) \pm 1.7}$ & ${6.9 \hspace{2pt} (\textcolor{red}{-1.3}) \pm 4.0}$ \\
    & MAProtoNet (ours) & ${86.5 \hspace{2pt} (\textcolor{green}{-1.7}) \pm 3.7}$ & ${\textbf{83.7} \hspace{2pt} (\textcolor{red}{+2.6}) \pm 2.9}$ & ${6.3 \hspace{2pt} (\textcolor{red}{-1.9}) \pm 3.5}$ \\
    % 2020
    \midrule
    \multirow{7}{*}{BraTS 2020} & CNN & ${85.5 \pm 4.6}_{\hspace{2pt} (86.5 \pm 2.6)}$ & ${10.7 \pm 5.7}_{\hspace{2pt} (9.9 \pm 3.0)}$ & ${13.8 \pm 7.8}_{\hspace{2pt} (11.2 \pm 4.9)}$ \\
    & ProtoPNet & ${84.3 \pm 2.7}_{\hspace{2pt} (86.8 \pm 3.2)}$ & ${11.8 \pm 2.8}_{\hspace{2pt} (0.8 \pm 0.3)}$ & ${24.2 \pm 7.7}_{\hspace{2pt} (60.9 \pm 16.4)}$ \\
    & XProtoNet & ${84.7 \pm 4.8}_{\hspace{2pt} (87.0 \pm 2.1)}$ & ${16.9 \pm 3.0}_{\hspace{2pt} (20.3 \pm 3.0)}$ & ${16.3 \pm 5.3}_{\hspace{2pt} (17.0 \pm 4.1)}$ \\
    & MProtoNet & ${85.8 \pm 1.9}_{\hspace{2pt} (85.8 \pm 4.8)}$ & ${81.2 \pm 3.7}_{(\hspace{2pt} 71.3 \pm 5.8)}$ & ${6.2 \pm 3.7}_{\hspace{2pt} (7.9 \pm 3.4)}$ \\
    & MProtoNet + $Q$ (ours) & ${85.8 \hspace{2pt} (\textcolor{black}{+0.0}) \pm 5.5}$ & ${81.5 \hspace{2pt} (\textcolor{red}{+0.3}) \pm 2.4}$ & ${6.4 \hspace{2pt} (\textcolor{green}{+0.2}) \pm 3.8}$ \\
    & MProtoNet + $MS$ (ours) & ${\textbf{87.6} \hspace{2pt} (\textcolor{red}{+1.8}) \pm 2.7}$ & ${79.0 \hspace{2pt} (\textcolor{green}{-2.2}) \pm 1.4}$ & ${8.6 \hspace{2pt} (\textcolor{green}{+1.6}) \pm 5.0}$ \\
    & MAProtoNet (ours) & ${86.7 \hspace{2pt} (\textcolor{red}{+0.9}) \pm 4.0}$ & ${\textbf{85.8} \hspace{2pt} (\textcolor{red}{+4.6}) \pm 2.0}$ & ${\textbf{6.2} \hspace{2pt} (\textcolor{black}{-0.0}) \pm 1.8}$ \\
    \bottomrule
  \end{tabular}
  \label{tab2}
\end{table*}

\subsection{Comparison of Baseline Methods}
We report our experimental results in Tab. \ref{tab2}. The results of our proposed method we report here are from MAProtoNet-c, which uses the architecture Fig. \ref{fig4:multi-scale}(c) for down-sampling and fusion. Further ablation study for different multi-scale modules are given in the following experiments.

As depicted in Tab. \ref{tab2}, overall, our MAProtoNet outperforms the baselines in terms of localization, while also preserving considerable classification and interpretability capabilities. Specifically, MAProtoNet improves AP score by about 2$\sim$4\% in all datasets as compared to MProtoNet, while maintaining similar BAC and IDS scores. 
Regarding AP score, our MAProtoNet exhibits an improvement of 4.6\% in BraTS 2020 dataset, but only showcases 2.6\% and 1.3\% improvement in BraTS 2019 and 2018 datasets. We attribute this discrepancy to the larger amount of training data available for BraTS 2020. Conversely, in BraTS 2019 and 2018, characterized by smaller datasets, the additional multi-scale module $MS$ and quadruplet attention $Q$ may lead to overfitting especially for the localization performance, thereby diminishing the score.

When focusing on the ablation results of our MAProtoNet, we note that quadruplet attention (denoted as MProtoNet + $Q$ in the table) consistently leads to a slight improvement in localization performance, whereas multi-scale module may usually lead to a decrease in AP score. Nevertheless, their combination yields substantially better results. We infer that solely introducing multi-scale features directly poses challenges for the model in learning to capture vital attribution information. Conversely, the attentive approach complements the multi-scale module by providing crucial clues as prior knowledge. Consequently, the combination of quadruplet attention and multi-scale features consistently yields superior performance. Details of complexity of the models are provided in \ref{app:complexity}.

\begin{figure*}[ht]
  \centering
  \includegraphics[width=0.95\textwidth]{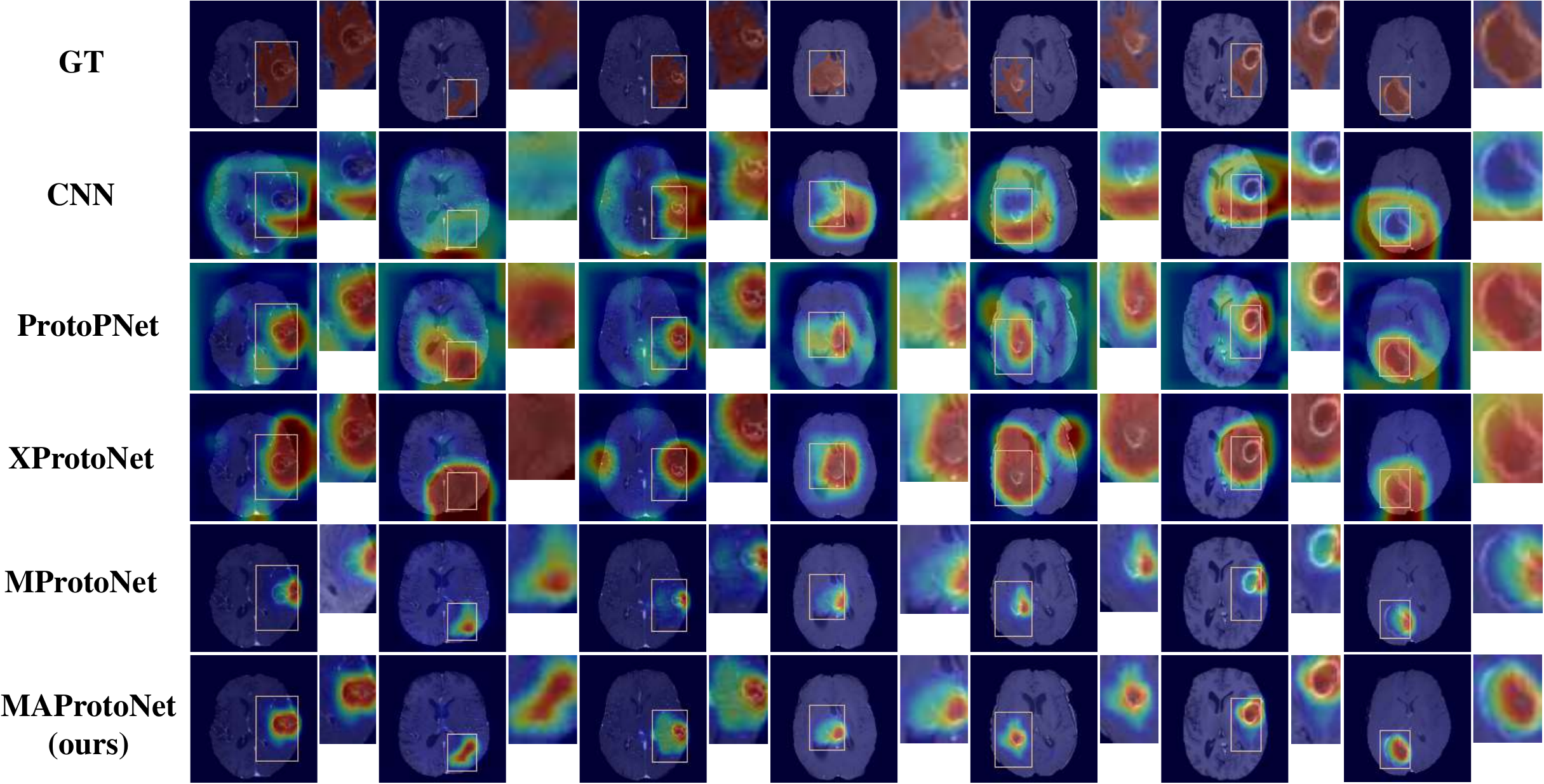}
  \caption{Visualization results of attribution maps. All examples here are from subjects in BraTS 2020 dataset. MRI slices of T1CE modality are reported. Our MAProtoNet showcases superior localization performance as compared to other baseline methods.}
  \label{fig4}
\end{figure*}

We visualize the ground truths and attribution maps in Fig. \ref{fig4}. The results demonstrate that our MAProtoNet exhibits stronger localization capability compared to the baseline models. More examples (both significant activation and non-significant activation cases) are provided in \ref{app:visual}.

\subsection{Ablation Study}
We conducted ablation experiments on architectures for multi-scale module as in Fig. \ref{fig4:multi-scale}(a)-(d), and the multi-scale mapping loss $\mathcal{L}_{mmap}$ as well, to validate their effectiveness.

\noindent
\textbf{Multi-scale Module.}
In order to explore the impact of different architectures to capture and fuse multi-scale features, we apply four designs shown in Fig. \ref{fig4:multi-scale} as the multi-scale module, comparing their performances in Tab. \ref{tab3}. 
Our observation indicates that MAProtoNet-c and MAProtoNet-d demonstrate relative stable performance in AP and IDS, owing to their utilization of both max pooling and average pooling layers, which can effectively incorporate strong prior knowledge and retain significant information during down-sampling with limited data. Furthermore, MAProtoNet-c, which concatenates features from all scales and fuses them using the following mapping module by channel augmentation, can achieve slightly better results as compared to MAProtoNet-d, the architecture that fuses features via learnable convolutional layers and addition operation in the multi-scale module directly. 
We deduce that the reason lies in the enhanced interactions across various granularity levels in MAProtoNet-c. Here, fusion takes place on augmented channels within the subsequent mapping module, rather than employing separate encoding with a convolutional layer beforehand as in MAProtoNet-d.
Taking these considerations into account, we employ architecture in Fig. \ref{fig4:multi-scale}(c) as our multi-scale module in all other experiments of the paper, and report the results of MAProtoNet-c to represent MAProtoNet. 

\begin{table}
  \centering
  \caption{Ablation results on different architectures for multi-scale module.}
  \begin{tabular}{c||c||ccc}
    \toprule
    Dataset & Method & BAC $\uparrow$ & AP $\uparrow$ & IDS $\downarrow$ \\
    % 2018
    \midrule
    \multirow{4}{*}{BraTS 2018} 
    & MAProtoNet-a & 86.4 & \textbf{83.1} & 5.9 \\
    & MAProtoNet-b & 84.9 & 80.8 & 4.7 \\
    & MAProtoNet-c & \textbf{87.7} & 82.2 & \textbf{4.1} \\
    & MAProtoNet-d & 86.2 & 82.4 & 11.1 \\
    % 2019
    \midrule
    \multirow{4}{*}{BraTS 2019}
    & MAProtoNet-a & \textbf{88.6} & 82.0 & 9.6 \\
    & MAProtoNet-b & 87.3 & 82.4 & 10.8 \\
    & MAProtoNet-c & 86.5 & \textbf{83.7} & \textbf{6.3} \\
    & MAProtoNet-d & 86.7 & 83.5 & 7.1 \\
    % 2020
    \midrule
    \multirow{4}{*}{BraTS 2020} 
    & MAProtoNet-a & $87.3$ & $85.0$ & $10.9$ \\
    & MAProtoNet-b & $84.7$ & $80.0$ & $8.6$ \\
    & MAProtoNet-c & $86.7$ & $\textbf{85.8}$ & $\textbf{6.2}$ \\
    & MAProtoNet-d & $\textbf{88.0}$ & $85.1$ & $7.9$ \\
    \bottomrule
  \end{tabular}
  \label{tab3}
\end{table}

\begin{table}
  \centering
  \caption{Ablation results on multi-scale mapping loss $\mathcal{L}_{mmap}$. w/o and w/ mean "without" and "with", respectively.}
  \begin{tabular}{c||c||ccc}
    \toprule
    Dataset & Method & BAC $\uparrow$ & AP $\uparrow$ & IDS $\downarrow$ \\
    % 2018
    \midrule
    \multirow{2}{*}{BraTS 2018} 
    & w/o $\mathcal{L}_{mmap}$ & 85.3 & 77.9 & 4.3 \\
    & w/ $\mathcal{L}_{mmap}$  & \textbf{87.7} & \textbf{82.2} & \textbf{4.1} \\
    % 2019
    \midrule
    \multirow{2}{*}{BraTS 2019}
    & w/o $\mathcal{L}_{mmap}$ & \textbf{88.7} & 79.5 & 16.7   \\
    & w/ $\mathcal{L}_{mmap}$ & 86.5 & \textbf{83.7} & \textbf{6.3} \\
    % 2020
    \midrule
    \multirow{2}{*}{BraTS 2020} 
    & w/o $\mathcal{L}_{mmap}$ & 86.1 & 81.9 & 7.8 \\
    & w/ $\mathcal{L}_{mmap}$ & \textbf{86.7} & \textbf{85.8}  & \textbf{6.2}  \\
    \bottomrule
  \end{tabular}
  \label{tab4}
\end{table}

\noindent
\textbf{Multi-scale Mapping Loss.} To investigate the role of the multi-scale mapping loss, we analyze the efficacy in Tab. \ref{tab4}. Specifically, we compare the performance of MAProtoNet with and without $\mathcal{L}_{mmap}$. We directly employ the model training with the loss function defined in Eq. \ref{eq12}, excluding the term of $\mathcal{L}_{mmap}$ as a control group. Results show that the MAProtoNet (with $\mathcal{L}_{mmap}$) can achieve much better performance in AP and IDS, proving that $\mathcal{L}_{mmap}$ does help improve the performance through additional supervision on both coarse-grained and fine-grained features against affine transformation.

\section{Conclusion and Future Works}
\label{conclusion}
In this work, we propose MAProtoNet, introducing a novel quadruplet attention and multi-scale module to improve the localization coherence of the attribution maps in existing interpretable  prototypical part network. Stemming from MProtoNet, the proposed quadruplet attention blocks can enhance the online CAM loss, while the multi-scale module captures and fuses both fine-grained and coarse-grained for better maps generation. 
Extensive experiments demonstrate that our MAProtoNet enhances the localization capability of the prototypical part network, achieving state-of-the-art performance in AP score with an increase of approximately 2\%$\sim$4\% in BraTS datasets, without significantly compromising classification performance.

To further advance, in the future, we can enhance our MAProtoNet in the following three aspects. Firstly, more unannotated data (without segmentation masks, only classification labels) should be employed for the network. In essence, MAProtoNet is a classification model, and its attribution capability is merely a byproduct. If we can leverage a large amount of cheaper classification data for training, its attribution generalization performance will be further enhanced. Secondly, more multi-scale information and more advanced multi-scale fusion architectures should be introduced. Taking into consideration the amount of training data and computational capacity, we set $n_{scales}=2$ and chose to employ a straightforward approach for fusion. If we can incorporate features from additional granularity levels and leverage more potent techniques, such as deformable attention and transformer decoder, for fusion, multi-scale information will be more effectively utilized. Thirdly, as large language models (LLMs) are famous for their capability to directly interact with users through natural language, collaborating LLMs with our MAProtoNet to produce a interaction pipeline is promising to enhance model interpretability.

\section*{Acknowledgments}
\label{acknowledgments}
We appreciate the authors of MProtoNet for their official source code \footnote{https://github.com/aywi/mprotonet}. We also appreciate the funding from FLOuRISH institute of Tokyo University of Agriculture and Technology (TUAT) \footnote{https://en.tuat-flourish.jp/}.

\appendix
\section{Data Pre-processing and Augmentation}
\label{app:data}
We apply a unified pipeline to pre-process and augment MRI data across all BraTS 2018, 2019, and 2020 datasets, ensuring consistency with the methods used in MProtoNet.
% MProtoNet
Specifically, each MRI image has a size of $240 \times 240 \times 155$ voxels, with resolution of $1 \times 1 \times 1$ mm$^3$. 
In data pre-processing, we crop the raw images into $192 \times 192 \times 192$, and then down-sample to $128 \times 128 \times 96$ voxels, with resolution of $1.5 \times 1.5 \times 1.5$ mm$^3$. Subsequently, we implement intensity normalization on each image by Z-score normalization. All four modalities: T1, T1c, T2 and FLAIR, are padding together onto channel dimension. 
In data augmentation, similar to the methods in nnU-Net, we employ a augmentations sequence of: (1) rotation and scaling; (2) Gaussian noise; (3) Gaussian blur; (4) brightness augmentation; (5) contrast augmentation; (6) simulation of low resolution; (7) gamma augmentation; (8) mirroring.
% Fabian Isensee, Paul F. Jaeger, Simon A. A. Kohl, Jens Petersen, and Klaus H. Maier-Hein. nnU-Net: A self-configuring method for deep learning-based biomedical image segmentation. Nature Methods, 18(2):203–211, February 2021. doi: 10.1038/s41592-020-01008-z.

\renewcommand{\thefigure}{E.1}
\begin{figure*}[htbp]
  \centering
  \includegraphics[width=0.95\textwidth]{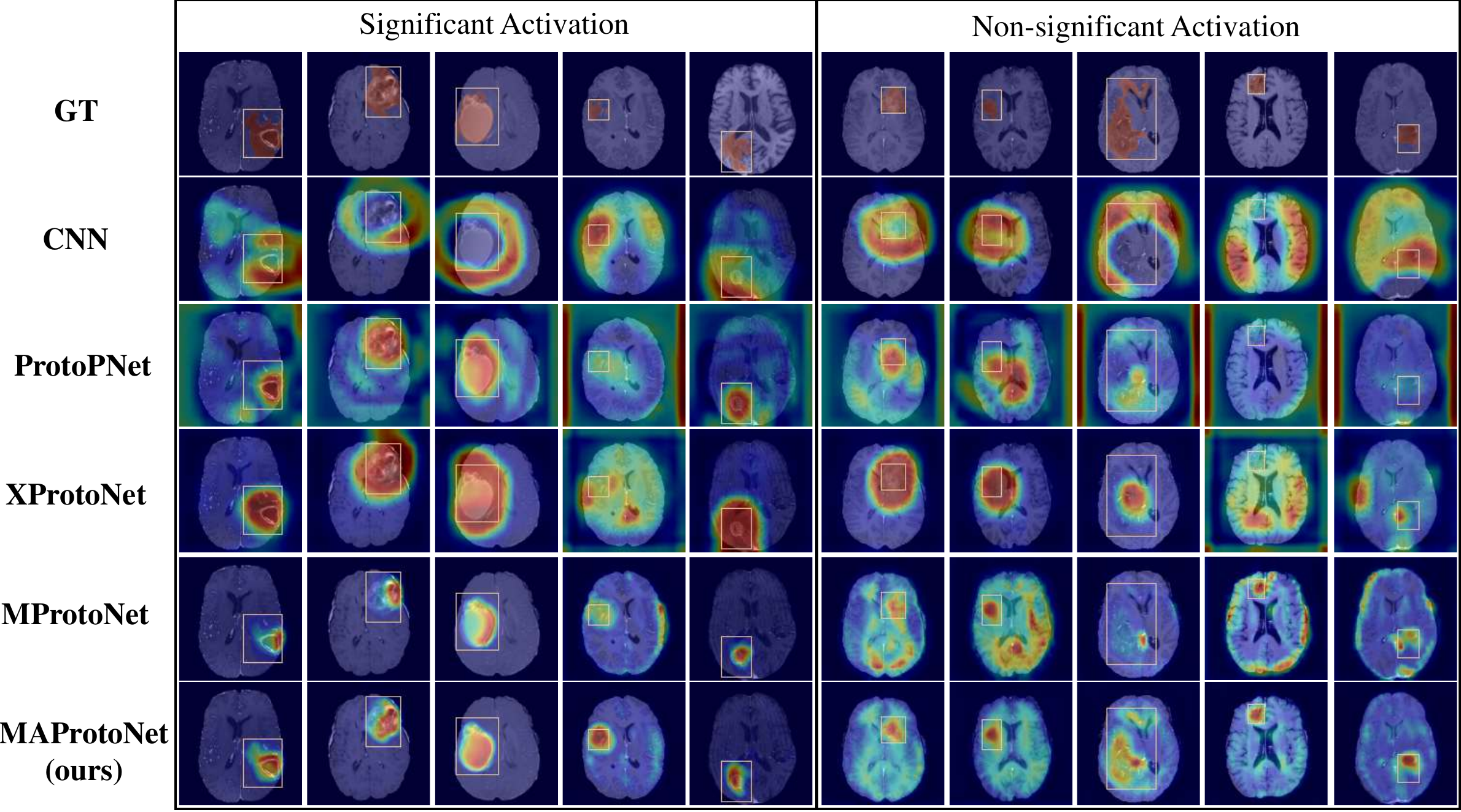}
  \caption{More visualization examples of attribution maps.}
  \label{fig:more}
\end{figure*}

\section{Evaluation Metrics}
\label{app:metrics}
\noindent
\textbf{Balanced Accuracy.} Given the sensitivity or true positive rate $TPR$, and specificity or true negative rate $TNR$ as well, BAC is defined as the average of $TPR$ and $TNR$:
\begin{align}
    BAC = \frac{TPR + TNR}{2} 
    \label{eq:bac}
\end{align}
The classification accuracy on the minority class contributes more to BAC. If the model performs equally on every class, BAC will reduce to conventional accuracy.
% Brodersen, K.H.; Ong, C.S.; Stephan, K.E.; Buhmann, J.M. (2010). The balanced accuracy and its posterior distribution. Proceedings of the 20th International Conference on Pattern Recognition, 3121-24.

\noindent
\textbf{Activation Precision.} Given the human-annotated WT masks $WT(x)$ and the resulting attribution map $M(x)$, AP can be defined as:
\begin{align}
    AP = \frac{WT(x) \cap T(\text{UpSample}(M(x)))}{T(\text{UpSample}(M(x)))} 
    \label{eq:ap}
\end{align}
where $T(\cdot)$ denotes the activation threshold, while UpSample$(\cdot)$ is the trilinear up-sampling operation. We follow previous work and set $T(\cdot)=0.5$. For XProtoNet, MProtoNet and our MAProtoNet, attribution maps are obtained by averaging on maps from the mapping module. 
% Alina Jade Barnett, Fides Regina Schwartz, Chaofan Tao, Chaofan Chen, Yinhao Ren, Joseph Y. Lo, and Cynthia Rudin. A case-based interpretable deep learning model for classification of mass lesions in digital mammography. Nature Machine Intelligence, 3 (12):1061–1070, December 2021. doi: 10.1038/s42256-021-00423-x.
For ProtoPNet, we use average of similarity scores on the top-$\alpha\%$ most activated patches as the attribution maps.
% Alina Jade Barnett, Fides Regina Schwartz, Chaofan Tao, Chaofan Chen, Yinhao Ren, Joseph Y. Lo, and Cynthia Rudin. A case-based interpretable deep learning model for classification of mass lesions in digital mammography. Nature Machine Intelligence, 3 (12):1061–1070, December 2021. doi: 10.1038/s42256-021-00423-x.
For CNN, we use the activated maps from the last convolution layer acquired via GradCAM for calculation.

\noindent
\textbf{Incremental Deletion Score.} Given the resulting attribution map with activated values ranging from 0 to 1, the incremental deletion curve is obtained by incrementally deleting features (e.g. replaced with 0) from the input, starting from high activated values to low. With the incremental deletion curve, IDS is defined as the normalized area under the curve within the bounds of start and end. IDS is commonly used to evaluate the interpretability of the resulting attribution map, by assessing whether the regions with high activation values are crucial for the classification task. Lower IDS indicates that the top-valued features contribute significantly, as removing these features would result in a rapid performance decline, thereby implying greater interpretability.
% Meike Nauta, Jan Trienes, Shreyasi Pathak, Elisa Nguyen, Michelle Peters, Yasmin Schmitt, J ̈org Schl ̈otterer, Maurice van Keulen, and Christin Seifert. From anecdotal evidence to quantitative evaluation methods: A systematic review on evaluating explainable AI. ACM Computing Surveys, February 2023. doi: 10.1145/3583558.
\section{Training Process}
\label{app:training}
The same as traditional prototypical part network, our MAProtoNet undergoes alternating training across three stages. 

\noindent
\textbf{Stage 1. Training on Models before classification layers.} We train the whole network except the classification module with loss defined in Eq. \ref{eq12}. 

\noindent
\textbf{Stage 2. Prototypes Reassignment.} We follow the thinking of case-based methods, replacing the learnable prototype vectors $v$, with features of a specific region in a training sample $map \odot fea$. The criterion for selection is the distance between the features and the learned vectors.

\noindent
\textbf{Stage 3. Adjusting.} We solely adjust the classification layer after reassignment, with cross entropy loss and the L1 normalization loss on the linear head.

The training sequences follow: Stage 1 $\to$ Stage 2 $\to$ Stage 3 $\to$ Stage 1 $\to$ ... $\to$ Stage 3. In our experiments, where the total number of epochs is set to 100, we alternate stages every 10 epochs. Specifically, we implement Stage 1 for 10 rounds, then update the prototype vectors, and adjust the classification layer exclusively for another 10 rounds. Subsequently, we repeat these steps for a total of 10 times.

\section{Model Complexity}
\label{app:complexity}
We demonstrate the model complexity in Tab. \ref{app:comp}. Modified from MProtoNet, our MAProtoNet-c achieves enhancement without a significant increase in model complexity.

\renewcommand{\thetable}{D.1}
\begin{table}[htbp]
  \centering
  \caption{Details of model complexity. We represent number of parameters as Params, and floating point operations per second as FLOPs.}
  \begin{tabular}{c||c||c}
    \toprule
    Model & Params $\downarrow$ & FLOPs $\downarrow$ \\
    \midrule
    MProtoNet & \textbf{0.61M} & \textbf{30.27G} \\
    MAProtoNet-a & 1.09M & 43.38G \\
    MAProtoNet-b & 1.06M & 42.57G \\
    MAProtoNet-c & 0.63M & 32.08G \\
    MAProtoNet-d & 0.65M & 32.51G \\
    \bottomrule
  \end{tabular}
  \label{app:comp}
\end{table}

\section{More Visualization Examples}
\label{app:visual}
We visualize more examples of the models in Fig. \ref{fig:more}. In both significant activation and non-significant activation cases, we can observe that our MAProtoNet consistently showcases superior localization abilities.





% References
% \input{references}
% \end{thebibliography}

\bibliographystyle{elsarticle-harv} 
\bibliography{MAProtoNet}

\begin{thebibliography}{61}
\expandafter\ifx\csname natexlab\endcsname\relax\def\natexlab#1{#1}\fi
\providecommand{\url}[1]{\texttt{#1}}
\providecommand{\href}[2]{#2}
\providecommand{\path}[1]{#1}
\providecommand{\DOIprefix}{doi:}
\providecommand{\ArXivprefix}{arXiv:}
\providecommand{\URLprefix}{URL: }
\providecommand{\Pubmedprefix}{pmid:}
\providecommand{\doi}[1]{\href{http://dx.doi.org/#1}{\path{#1}}}
\providecommand{\Pubmed}[1]{\href{pmid:#1}{\path{#1}}}
\providecommand{\bibinfo}[2]{#2}
\ifx\xfnm\relax \def\xfnm[#1]{\unskip,\space#1}\fi
%Type = Inproceedings
\bibitem[{Bau et~al.(2017)Bau, Zhou, Khosla, Oliva and Torralba}]{bau_network_2017}
\bibinfo{author}{Bau, D.}, \bibinfo{author}{Zhou, B.}, \bibinfo{author}{Khosla, A.}, \bibinfo{author}{Oliva, A.}, \bibinfo{author}{Torralba, A.}, \bibinfo{year}{2017}.
\newblock \bibinfo{title}{Network dissection: Quantifying interpretability of deep visual representations}, in: \bibinfo{booktitle}{Proceedings of the IEEE/CVF Conference on Computer Vision and Pattern Recognition (CVPR)}, \bibinfo{publisher}{{IEEE}}. pp. \bibinfo{pages}{3319--3327}.
\newblock \URLprefix \url{https://openaccess.thecvf.com/content_cvpr_2017/html/Bau_Network_Dissection_Quantifying_CVPR_2017_paper.html}, \DOIprefix\doi{10.1109/CVPR.2017.354}.
%Type = Article
\bibitem[{Bie et~al.(2024)Bie, Luo and Chen}]{bie_mica_2024}
\bibinfo{author}{Bie, Y.}, \bibinfo{author}{Luo, L.}, \bibinfo{author}{Chen, H.}, \bibinfo{year}{2024}.
\newblock \bibinfo{title}{{MICA}: Towards explainable skin lesion diagnosis via multi-level image-concept alignment}.
\newblock \bibinfo{journal}{arXiv preprint arXiv:2401.08527} \URLprefix \url{https://arxiv.org/abs/2401.08527}.
%Type = Article
\bibitem[{Camalan et~al.(2021)Camalan, Mahmood, Binol, Araújo, Santos-Silva, Vargas, Lopes, Khurram and Gurcan}]{cancers13061291}
\bibinfo{author}{Camalan, S.}, \bibinfo{author}{Mahmood, H.}, \bibinfo{author}{Binol, H.}, \bibinfo{author}{Araújo, A.L.D.}, \bibinfo{author}{Santos-Silva, A.R.}, \bibinfo{author}{Vargas, P.A.}, \bibinfo{author}{Lopes, M.A.}, \bibinfo{author}{Khurram, S.A.}, \bibinfo{author}{Gurcan, M.N.}, \bibinfo{year}{2021}.
\newblock \bibinfo{title}{Convolutional neural network-based clinical predictors of oral dysplasia: Class activation map analysis of deep learning results}.
\newblock \bibinfo{journal}{Cancers} \bibinfo{volume}{13}.
\newblock \URLprefix \url{https://www.mdpi.com/2072-6694/13/6/1291}, \DOIprefix\doi{10.3390/cancers13061291}.
%Type = Inproceedings
\bibitem[{Carloni et~al.(2023)Carloni, Berti, Iacconi, Pascali and Colantonio}]{rousseau_applicability_2023}
\bibinfo{author}{Carloni, G.}, \bibinfo{author}{Berti, A.}, \bibinfo{author}{Iacconi, C.}, \bibinfo{author}{Pascali, M.A.}, \bibinfo{author}{Colantonio, S.}, \bibinfo{year}{2023}.
\newblock \bibinfo{title}{On the applicability of prototypical part learning in medical images: Breast masses classification using {ProtoPNet}}, in: \bibinfo{booktitle}{Pattern Recognition, Computer Vision, and Image Processing. ICPR 2022 International Workshops and Challenges}, \bibinfo{publisher}{Springer Nature Switzerland}. pp. \bibinfo{pages}{539--557}.
\newblock \URLprefix \url{https://link.springer.com/10.1007/978-3-031-37660-3_38}, \DOIprefix\doi{10.1007/978-3-031-37660-3_38}.
%Type = Inproceedings
\bibitem[{Chen et~al.(2019)Chen, Li, Tao, Barnett, Rudin and Su}]{chen_this_2019}
\bibinfo{author}{Chen, C.}, \bibinfo{author}{Li, O.}, \bibinfo{author}{Tao, D.}, \bibinfo{author}{Barnett, A.}, \bibinfo{author}{Rudin, C.}, \bibinfo{author}{Su, J.K.}, \bibinfo{year}{2019}.
\newblock \bibinfo{title}{This looks like that: Deep learning for interpretable image recognition}, in: \bibinfo{booktitle}{Advances in Neural Information Processing Systems (NeurIPS)}, \bibinfo{publisher}{Curran Associates, Inc.}. pp. \bibinfo{pages}{8928--8939}.
\newblock \URLprefix \url{https://proceedings.neurips.cc/paper_files/paper/2019/file/adf7ee2dcf142b0e11888e72b43fcb75-Paper.pdf}.
%Type = Inproceedings
\bibitem[{Cheng et~al.(2022)Cheng, Misra, Schwing, Kirillov and Girdhar}]{cheng_masked-attention_2022}
\bibinfo{author}{Cheng, B.}, \bibinfo{author}{Misra, I.}, \bibinfo{author}{Schwing, A.G.}, \bibinfo{author}{Kirillov, A.}, \bibinfo{author}{Girdhar, R.}, \bibinfo{year}{2022}.
\newblock \bibinfo{title}{Masked-attention mask transformer for universal image segmentation}, in: \bibinfo{booktitle}{Proceedings of the IEEE/CVF Conference on Computer Vision and Pattern Recognition (CVPR)}, \bibinfo{publisher}{{IEEE}}. pp. \bibinfo{pages}{1290--1299}.
\newblock \URLprefix \url{https://openaccess.thecvf.com/content/CVPR2022/html/Cheng_Masked-Attention_Mask_Transformer_for_Universal_Image_Segmentation_CVPR_2022_paper.html}, \DOIprefix\doi{10.1109/CVPR52688.2022.00135}.
%Type = Inproceedings
\bibitem[{Cheng et~al.(2021)Cheng, Schwing and Kirillov}]{cheng_per-pixel_2021}
\bibinfo{author}{Cheng, B.}, \bibinfo{author}{Schwing, A.}, \bibinfo{author}{Kirillov, A.}, \bibinfo{year}{2021}.
\newblock \bibinfo{title}{Per-pixel classification is not all you need for semantic segmentation}, in: \bibinfo{booktitle}{Advances in Neural Information Processing Systems (NeurIPS)}, \bibinfo{publisher}{Curran Associates, Inc.}. pp. \bibinfo{pages}{17864--17875}.
\newblock \URLprefix \url{https://proceedings.neurips.cc/paper_files/paper/2021/file/950a4152c2b4aa3ad78bdd6b366cc179-Paper.pdf}.
%Type = Inproceedings
\bibitem[{Chowdhury et~al.(2021)Chowdhury, Santamaria-Pang, Kubricht and Tu}]{chowdhury_emergent_2021}
\bibinfo{author}{Chowdhury, A.}, \bibinfo{author}{Santamaria-Pang, A.}, \bibinfo{author}{Kubricht, J.R.}, \bibinfo{author}{Tu, P.}, \bibinfo{year}{2021}.
\newblock \bibinfo{title}{Emergent symbolic language based deep medical image classification}, in: \bibinfo{booktitle}{Proceedings of the 18th International Symposium on Biomedical Imaging (ISBI)}, \bibinfo{publisher}{{IEEE}}. pp. \bibinfo{pages}{689--692}.
\newblock \URLprefix \url{https://ieeexplore.ieee.org/document/9434073/}, \DOIprefix\doi{10.1109/ISBI48211.2021.9434073}.
%Type = Inproceedings
\bibitem[{Donnelly et~al.(2022)Donnelly, Barnett and Chen}]{donnelly_deformable_2022}
\bibinfo{author}{Donnelly, J.}, \bibinfo{author}{Barnett, A.J.}, \bibinfo{author}{Chen, C.}, \bibinfo{year}{2022}.
\newblock \bibinfo{title}{Deformable protopnet: An interpretable image classifier using deformable prototypes}, in: \bibinfo{booktitle}{Proceedings of the IEEE/CVF Conference on Computer Vision and Pattern Recognition (CVPR)}, \bibinfo{publisher}{{IEEE}}. pp. \bibinfo{pages}{10265--10275}.
\newblock \URLprefix \url{https://ieeexplore.ieee.org/document/9878975/}, \DOIprefix\doi{10.1109/CVPR52688.2022.01002}.
%Type = Inproceedings
\bibitem[{Dosovitskiy et~al.(2021)Dosovitskiy, Beyer, Kolesnikov, Weissenborn, Zhai, Unterthiner, Dehghani, Minderer, Heigold, Gelly, Uszkoreit and Houlsby}]{dosovitskiy_image_2021}
\bibinfo{author}{Dosovitskiy, A.}, \bibinfo{author}{Beyer, L.}, \bibinfo{author}{Kolesnikov, A.}, \bibinfo{author}{Weissenborn, D.}, \bibinfo{author}{Zhai, X.}, \bibinfo{author}{Unterthiner, T.}, \bibinfo{author}{Dehghani, M.}, \bibinfo{author}{Minderer, M.}, \bibinfo{author}{Heigold, G.}, \bibinfo{author}{Gelly, S.}, \bibinfo{author}{Uszkoreit, J.}, \bibinfo{author}{Houlsby, N.}, \bibinfo{year}{2021}.
\newblock \bibinfo{title}{An image is worth 16x16 words: Transformers for image recognition at scale}, in: \bibinfo{booktitle}{Proceedings of the 9th International Conference on Learning Representations ({ICLR})}, \bibinfo{publisher}{OpenReview.net}.
%Type = Article
\bibitem[{Geirhos et~al.(2018)Geirhos, Rubisch, Michaelis, Bethge, Wichmann and Brendel}]{geirhos_imagenet-trained_2022}
\bibinfo{author}{Geirhos, R.}, \bibinfo{author}{Rubisch, P.}, \bibinfo{author}{Michaelis, C.}, \bibinfo{author}{Bethge, M.}, \bibinfo{author}{Wichmann, F.A.}, \bibinfo{author}{Brendel, W.}, \bibinfo{year}{2018}.
\newblock \bibinfo{title}{{ImageNet}-trained {CNNs} are biased towards texture; increasing shape bias improves accuracy and robustness}.
\newblock \bibinfo{journal}{arXiv preprint arXiv:1811.12231} \URLprefix \url{http://arxiv.org/abs/1811.12231}.
%Type = Article
\bibitem[{Han et~al.(2021)Han, Chen, Tang, Lin, Jaiswal, Wang, Tewfik, Shih, Ding and Peng}]{han_using_2021}
\bibinfo{author}{Han, Y.}, \bibinfo{author}{Chen, C.}, \bibinfo{author}{Tang, L.}, \bibinfo{author}{Lin, M.}, \bibinfo{author}{Jaiswal, A.}, \bibinfo{author}{Wang, S.}, \bibinfo{author}{Tewfik, A.}, \bibinfo{author}{Shih, G.}, \bibinfo{author}{Ding, Y.}, \bibinfo{author}{Peng, Y.}, \bibinfo{year}{2021}.
\newblock \bibinfo{title}{Using radiomics as prior knowledge for thorax disease classification and localization in chest x-rays}.
\newblock \bibinfo{journal}{{AMIA} Annual Symposium Proceedings} \bibinfo{volume}{2021}, \bibinfo{pages}{546}.
\newblock \URLprefix \url{https://www.ncbi.nlm.nih.gov/pmc/articles/PMC8861661/}.
%Type = Inproceedings
\bibitem[{He et~al.(2016)He, Zhang, Ren and Sun}]{he_deep_2016}
\bibinfo{author}{He, K.}, \bibinfo{author}{Zhang, X.}, \bibinfo{author}{Ren, S.}, \bibinfo{author}{Sun, J.}, \bibinfo{year}{2016}.
\newblock \bibinfo{title}{Deep residual learning for image recognition}, in: \bibinfo{booktitle}{Proceedings of the IEEE/CVF Conference on Computer Vision and Pattern Recognition (CVPR)}, \bibinfo{publisher}{{IEEE}}. pp. \bibinfo{pages}{770--778}.
\newblock \URLprefix \url{https://openaccess.thecvf.com/content_cvpr_2016/html/He_Deep_Residual_Learning_CVPR_2016_paper.html}, \DOIprefix\doi{10.1109/CVPR.2016.90}.
%Type = Inproceedings
\bibitem[{Ho et~al.(2020)Ho, Jain and Abbeel}]{ho_denoising_2020}
\bibinfo{author}{Ho, J.}, \bibinfo{author}{Jain, A.}, \bibinfo{author}{Abbeel, P.}, \bibinfo{year}{2020}.
\newblock \bibinfo{title}{Denoising diffusion probabilistic models}, in: \bibinfo{booktitle}{Advances in Neural Information Processing Systems (NeurIPS)}, \bibinfo{publisher}{Curran Associates, Inc.}. pp. \bibinfo{pages}{6840--6851}.
\newblock \URLprefix \url{https://proceedings.neurips.cc/paper_files/paper/2020/file/4c5bcfec8584af0d967f1ab10179ca4b-Paper.pdf}.
%Type = Inproceedings
\bibitem[{Hu et~al.(2018)Hu, Shen and Sun}]{hu_squeeze-and-excitation_2019}
\bibinfo{author}{Hu, J.}, \bibinfo{author}{Shen, L.}, \bibinfo{author}{Sun, G.}, \bibinfo{year}{2018}.
\newblock \bibinfo{title}{Squeeze-and-excitation networks}, in: \bibinfo{booktitle}{Proceedings of the IEEE/CVF Conference on Computer Vision and Pattern Recognition (CVPR)}, \bibinfo{publisher}{{IEEE}}. pp. \bibinfo{pages}{7132--7141}.
\newblock \URLprefix \url{http://arxiv.org/abs/1709.01507}.
%Type = Inproceedings
\bibitem[{Huang et~al.(2020)Huang, Lin, Tong, Hu, Zhang, Iwamoto, Han, Chen and Wu}]{huang_unet_2020}
\bibinfo{author}{Huang, H.}, \bibinfo{author}{Lin, L.}, \bibinfo{author}{Tong, R.}, \bibinfo{author}{Hu, H.}, \bibinfo{author}{Zhang, Q.}, \bibinfo{author}{Iwamoto, Y.}, \bibinfo{author}{Han, X.}, \bibinfo{author}{Chen, Y.W.}, \bibinfo{author}{Wu, J.}, \bibinfo{year}{2020}.
\newblock \bibinfo{title}{{UNet} 3+: A full-scale connected {UNet} for medical image segmentation}, in: \bibinfo{booktitle}{{IEEE} International Conference on Acoustics, Speech and Signal Processing (ICASSP)}, \bibinfo{publisher}{{IEEE}}. pp. \bibinfo{pages}{1055--1059}.
\newblock \URLprefix \url{http://arxiv.org/abs/2004.08790}, \DOIprefix\doi{10.1109/ICASSP40776.2020.9053405}.
%Type = Inproceedings
\bibitem[{Ioffe and Szegedy(2015)}]{ioffe_batch_2015}
\bibinfo{author}{Ioffe, S.}, \bibinfo{author}{Szegedy, C.}, \bibinfo{year}{2015}.
\newblock \bibinfo{title}{Batch normalization: Accelerating deep network training by reducing internal covariate shift}, in: \bibinfo{booktitle}{Proceedings of the 32nd International Conference on Machine Learning (ICML)}, \bibinfo{publisher}{PMLR}. pp. \bibinfo{pages}{448--456}.
\newblock \URLprefix \url{https://proceedings.mlr.press/v37/ioffe15.html}.
%Type = Incollection
\bibitem[{Izadyyazdanabadi et~al.(2018)Izadyyazdanabadi, Belykh, Cavallo, Zhao, Gandhi, Moreira, Eschbacher, Nakaji, Preul and Yang}]{frangi_weakly-supervised_2018}
\bibinfo{author}{Izadyyazdanabadi, M.}, \bibinfo{author}{Belykh, E.}, \bibinfo{author}{Cavallo, C.}, \bibinfo{author}{Zhao, X.}, \bibinfo{author}{Gandhi, S.}, \bibinfo{author}{Moreira, L.B.}, \bibinfo{author}{Eschbacher, J.}, \bibinfo{author}{Nakaji, P.}, \bibinfo{author}{Preul, M.C.}, \bibinfo{author}{Yang, Y.}, \bibinfo{year}{2018}.
\newblock \bibinfo{title}{Weakly-supervised learning-based feature localization for confocal laser endomicroscopy glioma images}, in: \bibinfo{booktitle}{Medical Image Computing and Computer Assisted Intervention (MICCAI)}. \bibinfo{publisher}{Springer International Publishing}. volume \bibinfo{volume}{11071}, pp. \bibinfo{pages}{300--308}.
\newblock \URLprefix \url{https://link.springer.com/10.1007/978-3-030-00934-2_34}, \DOIprefix\doi{10.1007/978-3-030-00934-2_34}.
%Type = Inproceedings
\bibitem[{Jaiswal et~al.(2021)Jaiswal, Li, Zander, Han, Rousseau, Peng and Ding}]{jaiswal_scalp_2021}
\bibinfo{author}{Jaiswal, A.}, \bibinfo{author}{Li, T.}, \bibinfo{author}{Zander, C.}, \bibinfo{author}{Han, Y.}, \bibinfo{author}{Rousseau, J.F.}, \bibinfo{author}{Peng, Y.}, \bibinfo{author}{Ding, Y.}, \bibinfo{year}{2021}.
\newblock \bibinfo{title}{{SCALP} - supervised contrastive learning for cardiopulmonary disease classification and localization in chest x-rays using patient metadata}, in: \bibinfo{booktitle}{International Conference on Data Mining (ICDM)}, \bibinfo{publisher}{{IEEE}}. pp. \bibinfo{pages}{1132--1137}.
\newblock \URLprefix \url{https://ieeexplore.ieee.org/abstract/document/9679107}, \DOIprefix\doi{10.1109/ICDM51629.2021.00134}.
%Type = Inproceedings
\bibitem[{Kim et~al.(2021)Kim, Kim, Seo and Yoon}]{kim_xprotonet_2021}
\bibinfo{author}{Kim, E.}, \bibinfo{author}{Kim, S.}, \bibinfo{author}{Seo, M.}, \bibinfo{author}{Yoon, S.}, \bibinfo{year}{2021}.
\newblock \bibinfo{title}{{XProtoNet}: Diagnosis in chest radiography with global and local explanations}, in: \bibinfo{booktitle}{Proceedings of the IEEE/CVF Conference on Computer Vision and Pattern Recognition (CVPR)}, \bibinfo{publisher}{{IEEE}}. pp. \bibinfo{pages}{15719--15728}.
\newblock \URLprefix \url{https://ieeexplore.ieee.org/document/9577909/}, \DOIprefix\doi{10.1109/CVPR46437.2021.01546}.
%Type = Inproceedings
\bibitem[{Kim et~al.(2023)Kim, Kim, Choi and Kim}]{kim_concept_2023}
\bibinfo{author}{Kim, I.}, \bibinfo{author}{Kim, J.}, \bibinfo{author}{Choi, J.}, \bibinfo{author}{Kim, H.J.}, \bibinfo{year}{2023}.
\newblock \bibinfo{title}{Concept bottleneck with visual concept filtering for explainable medical image classification}, in: \bibinfo{booktitle}{Medical Image Computing and Computer Assisted Intervention (MICCAI) Workshops}, \bibinfo{publisher}{Springer Nature Switzerland}. pp. \bibinfo{pages}{225--233}.
\newblock \URLprefix \url{http://arxiv.org/abs/2308.11920}.
%Type = Article
\bibitem[{Kim et~al.(2019)Kim, Rajaraman and Antani}]{kim_visual_2019}
\bibinfo{author}{Kim, I.}, \bibinfo{author}{Rajaraman, S.}, \bibinfo{author}{Antani, S.}, \bibinfo{year}{2019}.
\newblock \bibinfo{title}{Visual interpretation of convolutional neural network predictions in classifying medical image modalities}.
\newblock \bibinfo{journal}{Diagnostics} \bibinfo{volume}{9}.
\newblock \URLprefix \url{https://www.mdpi.com/2075-4418/9/2/38}, \DOIprefix\doi{10.3390/diagnostics9020038}.
%Type = Inproceedings
\bibitem[{Koh et~al.(2020)Koh, Nguyen, Tang, Mussmann, Pierson, Kim and Liang}]{koh_concept_2020}
\bibinfo{author}{Koh, P.W.}, \bibinfo{author}{Nguyen, T.}, \bibinfo{author}{Tang, Y.S.}, \bibinfo{author}{Mussmann, S.}, \bibinfo{author}{Pierson, E.}, \bibinfo{author}{Kim, B.}, \bibinfo{author}{Liang, P.}, \bibinfo{year}{2020}.
\newblock \bibinfo{title}{Concept bottleneck models}, in: \bibinfo{booktitle}{Proceedings of the 37th International Conference on Machine Learning (ICML)}, \bibinfo{publisher}{PMLR}. pp. \bibinfo{pages}{5338--5348}.
\newblock \URLprefix \url{https://proceedings.mlr.press/v119/koh20a.html}.
%Type = Article
\bibitem[{Leming et~al.(2023)Leming, Bron, Bruffaerts, Ou, Iglesias, Gollub and Im}]{leming_challenges_2023}
\bibinfo{author}{Leming, M.J.}, \bibinfo{author}{Bron, E.E.}, \bibinfo{author}{Bruffaerts, R.}, \bibinfo{author}{Ou, Y.}, \bibinfo{author}{Iglesias, J.E.}, \bibinfo{author}{Gollub, R.L.}, \bibinfo{author}{Im, H.}, \bibinfo{year}{2023}.
\newblock \bibinfo{title}{Challenges of implementing computer-aided diagnostic models for neuroimages in a clinical setting}.
\newblock \bibinfo{journal}{npj Digital Medicine} \bibinfo{volume}{6}, \bibinfo{pages}{129}.
\newblock \URLprefix \url{https://doi.org/10.1038/s41746-023-00868-x}, \DOIprefix\doi{10.1038/s41746-023-00868-x}.
%Type = Article
\bibitem[{Li et~al.(2018)Li, Chen, Qi, Dou, Fu and Heng}]{li_h-denseunet_2018}
\bibinfo{author}{Li, X.}, \bibinfo{author}{Chen, H.}, \bibinfo{author}{Qi, X.}, \bibinfo{author}{Dou, Q.}, \bibinfo{author}{Fu, C.W.}, \bibinfo{author}{Heng, P.A.}, \bibinfo{year}{2018}.
\newblock \bibinfo{title}{H-{DenseUNet}: Hybrid densely connected {UNet} for liver and tumor segmentation from {CT} volumes}.
\newblock \bibinfo{journal}{{IEEE} Transactions on Medical Imaging} \bibinfo{volume}{37}, \bibinfo{pages}{2663--2674}.
\newblock \URLprefix \url{https://ieeexplore.ieee.org/document/8379359/}, \DOIprefix\doi{10.1109/TMI.2018.2845918}.
%Type = Article
\bibitem[{Lin et~al.(2022)Lin, Niu, Sui, Zhao, Zhuo and Calhoun}]{lin_sspnet_2022}
\bibinfo{author}{Lin, Q.H.}, \bibinfo{author}{Niu, Y.W.}, \bibinfo{author}{Sui, J.}, \bibinfo{author}{Zhao, W.D.}, \bibinfo{author}{Zhuo, C.}, \bibinfo{author}{Calhoun, V.D.}, \bibinfo{year}{2022}.
\newblock \bibinfo{title}{{SSPNet}: An interpretable 3d-{CNN} for classification of schizophrenia using phase maps of resting-state complex-valued {fMRI} data}.
\newblock \bibinfo{journal}{Medical Image Analysis} \bibinfo{volume}{79}, \bibinfo{pages}{102430}.
\newblock \URLprefix \url{https://www.sciencedirect.com/science/article/pii/S1361841522000810}, \DOIprefix\doi{https://doi.org/10.1016/j.media.2022.102430}.
%Type = Article
\bibitem[{Minaee et~al.(2022)Minaee, Boykov, Porikli, Plaza, Kehtarnavaz and Terzopoulos}]{minaee_image_2021}
\bibinfo{author}{Minaee, S.}, \bibinfo{author}{Boykov, Y.}, \bibinfo{author}{Porikli, F.}, \bibinfo{author}{Plaza, A.}, \bibinfo{author}{Kehtarnavaz, N.}, \bibinfo{author}{Terzopoulos, D.}, \bibinfo{year}{2022}.
\newblock \bibinfo{title}{Image segmentation using deep learning: A survey}.
\newblock \bibinfo{journal}{{IEEE} Transactions on Pattern Analysis and Machine Intelligence} \bibinfo{volume}{44}, \bibinfo{pages}{3523--3542}.
\newblock \URLprefix \url{https://ieeexplore.ieee.org/document/9356353/}, \DOIprefix\doi{10.1109/TPAMI.2021.3059968}.
%Type = Inproceedings
\bibitem[{Misra et~al.(2021)Misra, Nalamada, Arasanipalai and Hou}]{misra_rotate_2021}
\bibinfo{author}{Misra, D.}, \bibinfo{author}{Nalamada, T.}, \bibinfo{author}{Arasanipalai, A.U.}, \bibinfo{author}{Hou, Q.}, \bibinfo{year}{2021}.
\newblock \bibinfo{title}{Rotate to attend: Convolutional triplet attention module}, in: \bibinfo{booktitle}{Proceedings of the IEEE/CVF Winter Conference on Applications of Computer Vision (WACV)}, \bibinfo{publisher}{{IEEE}}. pp. \bibinfo{pages}{3139--3148}.
\newblock \URLprefix \url{https://ieeexplore.ieee.org/document/9423300/}, \DOIprefix\doi{10.1109/WACV48630.2021.00318}.
%Type = Inproceedings
\bibitem[{Mohammadjafari et~al.(2021)Mohammadjafari, Cevik, Thanabalasingam and Basar}]{mohammadjafari_using_2021}
\bibinfo{author}{Mohammadjafari, S.}, \bibinfo{author}{Cevik, M.}, \bibinfo{author}{Thanabalasingam, M.}, \bibinfo{author}{Basar, A.}, \bibinfo{year}{2021}.
\newblock \bibinfo{title}{Using {ProtoPNet} for interpretable alzheimer’s disease classification}, in: \bibinfo{booktitle}{Canadian Conference on Artificial Intelligence}.
\newblock \URLprefix \url{https://caiac.pubpub.org/pub/klwhoig4}, \DOIprefix\doi{10.21428/594757db.fb59ce6c}.
%Type = Inproceedings
\bibitem[{Nair and Hinton(2010)}]{nair_rectified_nodate}
\bibinfo{author}{Nair, V.}, \bibinfo{author}{Hinton, G.E.}, \bibinfo{year}{2010}.
\newblock \bibinfo{title}{Rectified linear units improve restricted boltzmann machines}, in: \bibinfo{booktitle}{Proceedings of the 27th International Conference on Machine Learning (ICML)}, \bibinfo{publisher}{Omnipress}. pp. \bibinfo{pages}{807--814}.
\newblock \URLprefix \url{https://www.cs.toronto.edu/~hinton/absps/reluICML.pdf}.
%Type = Article
\bibitem[{Natekar et~al.(2020)Natekar, Kori and Krishnamurthi}]{natekar_demystifying_2020}
\bibinfo{author}{Natekar, P.}, \bibinfo{author}{Kori, A.}, \bibinfo{author}{Krishnamurthi, G.}, \bibinfo{year}{2020}.
\newblock \bibinfo{title}{Demystifying brain tumor segmentation networks: Interpretability and uncertainty analysis}.
\newblock \bibinfo{journal}{Frontiers in Computational Neuroscience} \bibinfo{volume}{14}, \bibinfo{pages}{6}.
\newblock \URLprefix \url{https://www.frontiersin.org/articles/10.3389/fncom.2020.00006}, \DOIprefix\doi{10.3389/fncom.2020.00006}.
%Type = Inproceedings
\bibitem[{Nauta et~al.(2021)Nauta, van Bree and Seifert}]{nauta_neural_2021}
\bibinfo{author}{Nauta, M.}, \bibinfo{author}{van Bree, R.}, \bibinfo{author}{Seifert, C.}, \bibinfo{year}{2021}.
\newblock \bibinfo{title}{Neural prototype trees for interpretable fine-grained image recognition}, in: \bibinfo{booktitle}{Proceedings of the IEEE/CVF Conference on Computer Vision and Pattern Recognition (CVPR)}, \bibinfo{publisher}{{IEEE}}. pp. \bibinfo{pages}{14933--14943}.
\newblock \URLprefix \url{https://ieeexplore.ieee.org/document/9577335/}, \DOIprefix\doi{10.1109/CVPR46437.2021.01469}.
%Type = Inproceedings
\bibitem[{Nauta et~al.(2023)Nauta, Schl\"otterer, van Keulen and Seifert}]{nauta_pip-net_2023}
\bibinfo{author}{Nauta, M.}, \bibinfo{author}{Schl\"otterer, J.}, \bibinfo{author}{van Keulen, M.}, \bibinfo{author}{Seifert, C.}, \bibinfo{year}{2023}.
\newblock \bibinfo{title}{{PIP-Net}: Patch-based intuitive prototypes for interpretable image classification}, in: \bibinfo{booktitle}{Proceedings of the IEEE/CVF Conference on Computer Vision and Pattern Recognition (CVPR)}, \bibinfo{publisher}{{IEEE}}. pp. \bibinfo{pages}{2744--2753}.
\newblock \URLprefix \url{https://ieeexplore.ieee.org/document/10204807/}, \DOIprefix\doi{10.1109/CVPR52729.2023.00269}.
%Type = Article
\bibitem[{Omeiza et~al.(2019)Omeiza, Speakman, Cintas and Weldermariam}]{omeiza_smooth_2019}
\bibinfo{author}{Omeiza, D.}, \bibinfo{author}{Speakman, S.}, \bibinfo{author}{Cintas, C.}, \bibinfo{author}{Weldermariam, K.}, \bibinfo{year}{2019}.
\newblock \bibinfo{title}{Smooth grad-{CAM}++: An enhanced inference level visualization technique for deep convolutional neural network models}.
\newblock \bibinfo{journal}{arXiv preprint arXiv:1908.01224} \URLprefix \url{http://arxiv.org/abs/1908.01224}.
%Type = Inproceedings
\bibitem[{Patr{\'\i}cio et~al.(2023)Patr{\'\i}cio, Neves and Teixeira}]{patricio_coherent_2023}
\bibinfo{author}{Patr{\'\i}cio, C.}, \bibinfo{author}{Neves, J.a.C.}, \bibinfo{author}{Teixeira, L.F.}, \bibinfo{year}{2023}.
\newblock \bibinfo{title}{Coherent concept-based explanations in medical image and its application to skin lesion diagnosis}, in: \bibinfo{booktitle}{Proceedings of the IEEE/CVF Conference on Computer Vision and Pattern Recognition (CVPR) Workshops}, \bibinfo{publisher}{{IEEE}}. pp. \bibinfo{pages}{3799--3808}.
\newblock \URLprefix \url{https://ieeexplore.ieee.org/document/10208381/}, \DOIprefix\doi{10.1109/CVPRW59228.2023.00394}.
%Type = Article
\bibitem[{Patr\'{\i}cio et~al.(2023)Patr\'{\i}cio, Neves and Teixeira}]{patricio_explainable_2024}
\bibinfo{author}{Patr\'{\i}cio, C.}, \bibinfo{author}{Neves, J.a.C.}, \bibinfo{author}{Teixeira, L.F.}, \bibinfo{year}{2023}.
\newblock \bibinfo{title}{Explainable deep learning methods in medical image classification: A survey}.
\newblock \bibinfo{journal}{{ACM} Computing Surveys} \bibinfo{volume}{56}, \bibinfo{pages}{1--41}.
\newblock \URLprefix \url{https://doi.org/10.1145/3625287}, \DOIprefix\doi{10.1145/3625287}.
%Type = Inproceedings
\bibitem[{Pereira et~al.(2018)Pereira, Meier, Alves, Reyes and Silva}]{pereira_automatic_2018}
\bibinfo{author}{Pereira, S.}, \bibinfo{author}{Meier, R.}, \bibinfo{author}{Alves, V.}, \bibinfo{author}{Reyes, M.}, \bibinfo{author}{Silva, C.A.}, \bibinfo{year}{2018}.
\newblock \bibinfo{title}{Automatic brain tumor grading from {MRI} data using convolutional neural networks and quality assessment}, in: \bibinfo{booktitle}{Understanding and Interpreting Machine Learning in Medical Image Computing Applications}, \bibinfo{publisher}{Springer International Publishing}. pp. \bibinfo{pages}{106--114}.
\newblock \URLprefix \url{http://arxiv.org/abs/1809.09468}, \DOIprefix\doi{10.1007/978-3-030-02628-8}.
%Type = Inproceedings
\bibitem[{Pisov et~al.(2019)Pisov, Goncharov, Kurochkina, Morozov, Gombolevskiy, Chernina, Vladzymyrskyy, Zamyatina, Chesnokova, Pronin et~al.}]{pisov_incorporating_2019}
\bibinfo{author}{Pisov, M.}, \bibinfo{author}{Goncharov, M.}, \bibinfo{author}{Kurochkina, N.}, \bibinfo{author}{Morozov, S.}, \bibinfo{author}{Gombolevskiy, V.}, \bibinfo{author}{Chernina, V.}, \bibinfo{author}{Vladzymyrskyy, A.}, \bibinfo{author}{Zamyatina, K.}, \bibinfo{author}{Chesnokova, A.}, \bibinfo{author}{Pronin, I.}, et~al., \bibinfo{year}{2019}.
\newblock \bibinfo{title}{Incorporating task-specific structural knowledge into {CNNs} for brain midline shift detection}, in: \bibinfo{booktitle}{Interpretability of Machine Intelligence in Medical Image Computing and Multimodal Learning for Clinical Decision Support}, \bibinfo{publisher}{{Springer}}. pp. \bibinfo{pages}{30--38}.
\newblock \URLprefix \url{http://arxiv.org/abs/1908.04568}.
%Type = Inproceedings
\bibitem[{Ronneberger et~al.(2015)Ronneberger, Fischer and Brox}]{ronneberger_u-net_2015}
\bibinfo{author}{Ronneberger, O.}, \bibinfo{author}{Fischer, P.}, \bibinfo{author}{Brox, T.}, \bibinfo{year}{2015}.
\newblock \bibinfo{title}{{U-Net}: Convolutional networks for biomedical image segmentation}, in: \bibinfo{booktitle}{Medical Image Computing and Computer Assisted Intervention (MICCAI)}, \bibinfo{publisher}{Springer International Publishing}. pp. \bibinfo{pages}{234--241}.
\newblock \URLprefix \url{http://arxiv.org/abs/1505.04597}.
%Type = Inproceedings
\bibitem[{Rymarczyk et~al.(2022)Rymarczyk, Struski, G{\'o}rszczak, Lewandowska, Tabor and Zieli{\'{n}}ski}]{rymarczyk_interpretable_2022}
\bibinfo{author}{Rymarczyk, D.}, \bibinfo{author}{Struski, {\L}.}, \bibinfo{author}{G{\'o}rszczak, M.}, \bibinfo{author}{Lewandowska, K.}, \bibinfo{author}{Tabor, J.}, \bibinfo{author}{Zieli{\'{n}}ski, B.}, \bibinfo{year}{2022}.
\newblock \bibinfo{title}{Interpretable image classification with differentiable prototypes assignment}, in: \bibinfo{booktitle}{Proceedings of the European Conference on Computer Vision (ECCV)}, \bibinfo{publisher}{Springer Nature Switzerland}. pp. \bibinfo{pages}{351--368}.
\newblock \URLprefix \url{http://arxiv.org/abs/2112.02902}.
%Type = Inproceedings
\bibitem[{Rymarczyk et~al.(2020)Rymarczyk, Struski, Tabor and Zieli{\'n}ski}]{rymarczyk_protopshare_2021}
\bibinfo{author}{Rymarczyk, D.}, \bibinfo{author}{Struski, {\L}.}, \bibinfo{author}{Tabor, J.}, \bibinfo{author}{Zieli{\'n}ski, B.}, \bibinfo{year}{2020}.
\newblock \bibinfo{title}{{ProtoPShare}: Prototype sharing for interpretable image classification and similarity discovery}, in: \bibinfo{booktitle}{Proceedings of the 27th {ACM} {SIGKDD} Conference on Knowledge Discovery \& Data Mining}, pp. \bibinfo{pages}{1420--1430}.
\newblock \URLprefix \url{http://arxiv.org/abs/2011.14340}, \DOIprefix\doi{10.1145/3447548.3467245}.
%Type = Inproceedings
\bibitem[{Sacha et~al.(2023)Sacha, Rymarczyk, Struski, Tabor and Zieli\'nski}]{sacha_protoseg_2023}
\bibinfo{author}{Sacha, M.}, \bibinfo{author}{Rymarczyk, D.}, \bibinfo{author}{Struski, {\L}.}, \bibinfo{author}{Tabor, J.}, \bibinfo{author}{Zieli\'nski, B.}, \bibinfo{year}{2023}.
\newblock \bibinfo{title}{{ProtoSeg}: Interpretable semantic segmentation with prototypical parts}, in: \bibinfo{booktitle}{Proceedings of the IEEE/CVF Winter Conference on Applications of Computer Vision (WACV)}, \bibinfo{publisher}{{IEEE}}. pp. \bibinfo{pages}{1481--1492}.
\newblock \URLprefix \url{https://ieeexplore.ieee.org/document/10030923/}, \DOIprefix\doi{10.1109/WACV56688.2023.00153}.
%Type = Article
\bibitem[{Salahuddin et~al.(2022)Salahuddin, Woodruff, Chatterjee and Lambin}]{salahuddin_transparency_2022}
\bibinfo{author}{Salahuddin, Z.}, \bibinfo{author}{Woodruff, H.C.}, \bibinfo{author}{Chatterjee, A.}, \bibinfo{author}{Lambin, P.}, \bibinfo{year}{2022}.
\newblock \bibinfo{title}{Transparency of deep neural networks for medical image analysis: A review of interpretability methods}.
\newblock \bibinfo{journal}{Computers in Biology and Medicine} \bibinfo{volume}{140}, \bibinfo{pages}{105111}.
\newblock \URLprefix \url{https://www.sciencedirect.com/science/article/pii/S0010482521009057}, \DOIprefix\doi{https://doi.org/10.1016/j.compbiomed.2021.105111}.
%Type = Inproceedings
\bibitem[{Santamaria-Pang et~al.(2020)Santamaria-Pang, Kubricht, Chowdhury, Bhushan and Tu}]{martel_towards_2020}
\bibinfo{author}{Santamaria-Pang, A.}, \bibinfo{author}{Kubricht, J.}, \bibinfo{author}{Chowdhury, A.}, \bibinfo{author}{Bhushan, C.}, \bibinfo{author}{Tu, P.}, \bibinfo{year}{2020}.
\newblock \bibinfo{title}{Towards emergent language symbolic semantic segmentation and model interpretability}, in: \bibinfo{booktitle}{Medical Image Computing and Computer Assisted Intervention {MICCAI}}, \bibinfo{publisher}{Springer}. pp. \bibinfo{pages}{326--334}.
\newblock \URLprefix \url{https://link.springer.com/10.1007/978-3-030-59710-8_32}, \DOIprefix\doi{10.1007/978-3-030-59710-8_32}.
%Type = Inproceedings
\bibitem[{Selvaraju et~al.(2017)Selvaraju, Cogswell, Das, Vedantam, Parikh and Batra}]{selvaraju_grad-cam_nodate}
\bibinfo{author}{Selvaraju, R.R.}, \bibinfo{author}{Cogswell, M.}, \bibinfo{author}{Das, A.}, \bibinfo{author}{Vedantam, R.}, \bibinfo{author}{Parikh, D.}, \bibinfo{author}{Batra, D.}, \bibinfo{year}{2017}.
\newblock \bibinfo{title}{{Grad-CAM}: Visual explanations from deep networks via gradient-based localization}, in: \bibinfo{booktitle}{Proceedings of the IEEE International Conference on Computer Vision (ICCV)}, \bibinfo{publisher}{{IEEE}}. pp. \bibinfo{pages}{618--626}.
\newblock \URLprefix \url{https://openaccess.thecvf.com/content_iccv_2017/html/Selvaraju_Grad-CAM_Visual_Explanations_ICCV_2017_paper.html}.
%Type = Article
\bibitem[{Shastry and Sanjay(2022)}]{shastry_cancer_2022}
\bibinfo{author}{Shastry, K.A.}, \bibinfo{author}{Sanjay, H.A.}, \bibinfo{year}{2022}.
\newblock \bibinfo{title}{Cancer diagnosis using artificial intelligence: A review}.
\newblock \bibinfo{journal}{Artificial Intelligence Review} \bibinfo{volume}{55}, \bibinfo{pages}{2641--2673}.
\newblock \URLprefix \url{https://doi.org/10.1007/s10462-021-10074-4}, \DOIprefix\doi{10.1007/s10462-021-10074-4}.
%Type = Article
\bibitem[{Srivastava et~al.(2014)Srivastava, Hinton, Krizhevsky, Sutskever and Salakhutdinov}]{srivastava_dropout_nodate}
\bibinfo{author}{Srivastava, N.}, \bibinfo{author}{Hinton, G.}, \bibinfo{author}{Krizhevsky, A.}, \bibinfo{author}{Sutskever, I.}, \bibinfo{author}{Salakhutdinov, R.}, \bibinfo{year}{2014}.
\newblock \bibinfo{title}{Dropout: A simple way to prevent neural networks from overﬁtting}.
\newblock \bibinfo{journal}{Journal of Machine Learning Research} \bibinfo{volume}{15}, \bibinfo{pages}{1929--1958}.
\newblock \URLprefix \url{https://www.cs.toronto.edu/~hinton/absps/JMLRdropout.pdf}.
%Type = Inproceedings
\bibitem[{Sun et~al.(2020)Sun, Darbehani, Zaidi and Wang}]{sun_saunet_2020}
\bibinfo{author}{Sun, J.}, \bibinfo{author}{Darbehani, F.}, \bibinfo{author}{Zaidi, M.}, \bibinfo{author}{Wang, B.}, \bibinfo{year}{2020}.
\newblock \bibinfo{title}{{SAUNet}: Shape attentive {U-Net} for interpretable medical image segmentation}, in: \bibinfo{booktitle}{Medical Image Computing and Computer Assisted Intervention (MICCAI)}, \bibinfo{publisher}{Springer}. pp. \bibinfo{pages}{797--806}.
\newblock \URLprefix \url{http://arxiv.org/abs/2001.07645}.
%Type = Article
\bibitem[{Temme(2017)}]{temme_algorithms_2017}
\bibinfo{author}{Temme, M.}, \bibinfo{year}{2017}.
\newblock \bibinfo{title}{Algorithms and transparency in view of the new general data protection regulation}.
\newblock \bibinfo{journal}{European Data Protection Law Review ({EDPL})} \bibinfo{volume}{3}, \bibinfo{pages}{473}.
\newblock \URLprefix \url{https://heinonline.org/HOL/Page?handle=hein.journals/edpl3&id=512&div=&collection=}.
%Type = Inproceedings
\bibitem[{Vaswani et~al.(2017)Vaswani, Shazeer, Parmar, Uszkoreit, Jones, Gomez, Kaiser and Polosukhin}]{vaswani_attention_2017}
\bibinfo{author}{Vaswani, A.}, \bibinfo{author}{Shazeer, N.}, \bibinfo{author}{Parmar, N.}, \bibinfo{author}{Uszkoreit, J.}, \bibinfo{author}{Jones, L.}, \bibinfo{author}{Gomez, A.N.}, \bibinfo{author}{Kaiser, L.u.}, \bibinfo{author}{Polosukhin, I.}, \bibinfo{year}{2017}.
\newblock \bibinfo{title}{Attention is all you need}, in: \bibinfo{booktitle}{Advances in Neural Information Processing Systems (NeurIPS)}, \bibinfo{publisher}{Curran Associates, Inc.}. pp. \bibinfo{pages}{5998--6008}.
\newblock \URLprefix \url{https://proceedings.neurips.cc/paper_files/paper/2017/file/3f5ee243547dee91fbd053c1c4a845aa-Paper.pdf}.
%Type = Inproceedings
\bibitem[{Wang et~al.(2022)Wang, Xu, Zhang, Wang, Zheng and Liu}]{wang_convolutional_2022}
\bibinfo{author}{Wang, C.}, \bibinfo{author}{Xu, H.}, \bibinfo{author}{Zhang, X.}, \bibinfo{author}{Wang, L.}, \bibinfo{author}{Zheng, Z.}, \bibinfo{author}{Liu, H.}, \bibinfo{year}{2022}.
\newblock \bibinfo{title}{Convolutional embedding makes hierarchical vision transformer stronger}, in: \bibinfo{booktitle}{Proceedings of the European Conference on Computer Vision (ECCV)}, \bibinfo{publisher}{Springer Nature Switzerland}. pp. \bibinfo{pages}{739--756}.
\newblock \URLprefix \url{https://www.ecva.net/papers/eccv_2022/papers_ECCV/html/3627_ECCV_2022_paper.php}.
%Type = Inproceedings
\bibitem[{Wang et~al.(2023)Wang, Jiajie, Liu and Zhao}]{wang_hqprotopnet_2023}
\bibinfo{author}{Wang, J.}, \bibinfo{author}{Jiajie, P.}, \bibinfo{author}{Liu, Z.}, \bibinfo{author}{Zhao, H.}, \bibinfo{year}{2023}.
\newblock \bibinfo{title}{{HQProtoPNet}: An evidence-based model for interpretable image recognition}, in: \bibinfo{booktitle}{International Joint Conference on Neural Networks (IJCNN)}, \bibinfo{publisher}{{IEEE}}. pp. \bibinfo{pages}{1--8}.
\newblock \URLprefix \url{https://ieeexplore.ieee.org/abstract/document/10191863}, \DOIprefix\doi{10.1109/IJCNN54540.2023.10191863}.
%Type = Inproceedings
\bibitem[{Wang et~al.(2018)Wang, Peng, Lu, Lu and Summers}]{wang_tienet_2018}
\bibinfo{author}{Wang, X.}, \bibinfo{author}{Peng, Y.}, \bibinfo{author}{Lu, L.}, \bibinfo{author}{Lu, Z.}, \bibinfo{author}{Summers, R.M.}, \bibinfo{year}{2018}.
\newblock \bibinfo{title}{Tienet: Text-image embedding network for common thorax disease classification and reporting in chest x-rays}, in: \bibinfo{booktitle}{Proceedings of the IEEE/CVF Conference on Computer Vision and Pattern Recognition (CVPR)}, \bibinfo{publisher}{{IEEE}}. pp. \bibinfo{pages}{9049--9058}.
\newblock \URLprefix \url{https://ieeexplore.ieee.org/document/8579041/}, \DOIprefix\doi{10.1109/CVPR.2018.00943}.
%Type = Inproceedings
\bibitem[{Wei et~al.(2024)Wei, Tam and Tang}]{wei_mprotonet_2024}
\bibinfo{author}{Wei, Y.}, \bibinfo{author}{Tam, R.}, \bibinfo{author}{Tang, X.}, \bibinfo{year}{2024}.
\newblock \bibinfo{title}{{MProtoNet}: A case-based interpretable model for brain tumor classification with 3d multi-parametric magnetic resonance imaging}, in: \bibinfo{booktitle}{Medical Imaging with Deep Learning (MIDL)}, \bibinfo{publisher}{PMLR}. pp. \bibinfo{pages}{1798--1812}.
\newblock \URLprefix \url{https://proceedings.mlr.press/v227/wei24a.html}.
%Type = Inproceedings
\bibitem[{Woo et~al.(2018)Woo, Park, Lee and Kweon}]{woo_cbam_2018}
\bibinfo{author}{Woo, S.}, \bibinfo{author}{Park, J.}, \bibinfo{author}{Lee, J.Y.}, \bibinfo{author}{Kweon, I.S.}, \bibinfo{year}{2018}.
\newblock \bibinfo{title}{{CBAM}: Convolutional block attention module}, in: \bibinfo{booktitle}{Proceedings of the European Conference on Computer Vision (ECCV)}, \bibinfo{publisher}{{IEEE}}. pp. \bibinfo{pages}{3--19}.
\newblock \URLprefix \url{http://arxiv.org/abs/1807.06521}.
%Type = Article
\bibitem[{Yan et~al.(2023)Yan, Wang, Zhong, He, Karypis, Wang, Dong, Gentili, Hsu, Shang et~al.}]{yan_robust_2023}
\bibinfo{author}{Yan, A.}, \bibinfo{author}{Wang, Y.}, \bibinfo{author}{Zhong, Y.}, \bibinfo{author}{He, Z.}, \bibinfo{author}{Karypis, P.}, \bibinfo{author}{Wang, Z.}, \bibinfo{author}{Dong, C.}, \bibinfo{author}{Gentili, A.}, \bibinfo{author}{Hsu, C.N.}, \bibinfo{author}{Shang, J.}, et~al., \bibinfo{year}{2023}.
\newblock \bibinfo{title}{Robust and interpretable medical image classifiers via concept bottleneck models}.
\newblock \bibinfo{journal}{arXiv preprint arXiv:2310.03182} \URLprefix \url{http://arxiv.org/abs/2310.03182}.
%Type = Inproceedings
\bibitem[{Yuan et~al.(2023)Yuan, Xia, Dong, Chen, Yao, Qiu, Yan, Yin, Shi, Chen, Liu, Dong, Zhou, Lu, Zhang and Zhang}]{yuan_devil_2023}
\bibinfo{author}{Yuan, M.}, \bibinfo{author}{Xia, Y.}, \bibinfo{author}{Dong, H.}, \bibinfo{author}{Chen, Z.}, \bibinfo{author}{Yao, J.}, \bibinfo{author}{Qiu, M.}, \bibinfo{author}{Yan, K.}, \bibinfo{author}{Yin, X.}, \bibinfo{author}{Shi, Y.}, \bibinfo{author}{Chen, X.}, \bibinfo{author}{Liu, Z.}, \bibinfo{author}{Dong, B.}, \bibinfo{author}{Zhou, J.}, \bibinfo{author}{Lu, L.}, \bibinfo{author}{Zhang, L.}, \bibinfo{author}{Zhang, L.}, \bibinfo{year}{2023}.
\newblock \bibinfo{title}{Devil is in the queries: Advancing mask transformers for real-world medical image segmentation and out-of-distribution localization}, in: \bibinfo{booktitle}{Proceedings of the IEEE/CVF Conference on Computer Vision and Pattern Recognition (CVPR)}, \bibinfo{publisher}{{IEEE}}. pp. \bibinfo{pages}{23879--23889}.
\newblock \URLprefix \url{https://ieeexplore.ieee.org/document/10203355/}, \DOIprefix\doi{10.1109/CVPR52729.2023.02287}.
%Type = Article
\bibitem[{Zhang et~al.(2019)Zhang, Chen, McGough, Xing, Wang, Bui, Xie, Sapkota, Cui, Dhillon et~al.}]{zhang_pathologist-level_2019}
\bibinfo{author}{Zhang, Z.}, \bibinfo{author}{Chen, P.}, \bibinfo{author}{McGough, M.}, \bibinfo{author}{Xing, F.}, \bibinfo{author}{Wang, C.}, \bibinfo{author}{Bui, M.}, \bibinfo{author}{Xie, Y.}, \bibinfo{author}{Sapkota, M.}, \bibinfo{author}{Cui, L.}, \bibinfo{author}{Dhillon, J.}, et~al., \bibinfo{year}{2019}.
\newblock \bibinfo{title}{Pathologist-level interpretable whole-slide cancer diagnosis with deep learning}.
\newblock \bibinfo{journal}{Nature Machine Intelligence} \bibinfo{volume}{1}, \bibinfo{pages}{236--245}.
\newblock \URLprefix \url{https://www.nature.com/articles/s42256-019-0052-1}, \DOIprefix\doi{10.1038/s42256-019-0052-1}.
%Type = Article
\bibitem[{Zhao(2023)}]{zhao_crosseai_2023}
\bibinfo{author}{Zhao, J.}, \bibinfo{year}{2023}.
\newblock \bibinfo{title}{{CrossEAI}: Using explainable {AI} to generate better bounding boxes for chest x-ray images}.
\newblock \bibinfo{journal}{arXiv preprint arXiv:2310.19835} \URLprefix \url{http://arxiv.org/abs/2310.19835}.
%Type = Inproceedings
\bibitem[{Zhou et~al.(2016)Zhou, Khosla, Lapedriza, Oliva and Torralba}]{zhou_learning_2016}
\bibinfo{author}{Zhou, B.}, \bibinfo{author}{Khosla, A.}, \bibinfo{author}{Lapedriza, A.}, \bibinfo{author}{Oliva, A.}, \bibinfo{author}{Torralba, A.}, \bibinfo{year}{2016}.
\newblock \bibinfo{title}{Learning deep features for discriminative localization}, in: \bibinfo{booktitle}{Proceedings of the IEEE/CVF Conference on Computer Vision and Pattern Recognition (CVPR)}, \bibinfo{publisher}{{IEEE}}. pp. \bibinfo{pages}{2921--2929}.
\newblock \URLprefix \url{http://ieeexplore.ieee.org/document/7780688/}, \DOIprefix\doi{10.1109/CVPR.2016.319}.
%Type = Inproceedings
\bibitem[{Zhu et~al.(2021)Zhu, Su, Lu, Li, Wang and Dai}]{zhu_deformable_2021}
\bibinfo{author}{Zhu, X.}, \bibinfo{author}{Su, W.}, \bibinfo{author}{Lu, L.}, \bibinfo{author}{Li, B.}, \bibinfo{author}{Wang, X.}, \bibinfo{author}{Dai, J.}, \bibinfo{year}{2021}.
\newblock \bibinfo{title}{Deformable {DETR}: Deformable transformers for end-to-end object detection}, in: \bibinfo{booktitle}{Proceedings of the 9th International Conference on Learning Representations (ICLR) 2021}, \bibinfo{publisher}{OpenReview.net}.
\newblock \URLprefix \url{https://openreview.net/forum?id=gZ9hCDWe6ke}.

\end{thebibliography}

\end{document}